\documentclass{article}

\usepackage{arxiv}

\usepackage[utf8]{inputenc} 
\usepackage[T1]{fontenc}    
\usepackage{hyperref}       
\usepackage{url}            
\usepackage{booktabs}       
\usepackage{amsfonts}       
\usepackage{nicefrac}       
\usepackage{microtype}      
\usepackage{lipsum}         
\usepackage{graphicx}
\usepackage{doi}
\usepackage{authblk}
\usepackage{adjustbox}
\usepackage{threeparttable}
\usepackage{longtable}
\usepackage{multirow}
\usepackage{tcolorbox}
\tcbuselibrary{skins, breakable}
\usepackage{listings}
\usepackage{array}
\usepackage{fancyvrb}
\usepackage{siunitx}
\usepackage{subcaption}
\usepackage[numbers,sort&compress]{natbib}
\usepackage{tabularx}

\title{Visuospatial Perspective Taking in Multimodal Language Models}

\author{
  \textbf{Jonathan Prunty}$^{1}$\thanks{Correspondence to: jep84@cam.ac.uk} \quad
  \textbf{Seraphina Zhang}$^{1,2}$ \quad
  \textbf{Patrick Quinn}$^{1}$ \quad
  \textbf{Jianxun Lian}$^{3}$ \quad
  \textbf{Xing Xie}$^{3}$ \quad
  \textbf{Lucy Cheke}$^{2}$ \\
  \vspace{0.4em}
  \small
  \begin{tabular}{@{}c@{}}
    $^{1}$Leverhulme Centre for the Future of Intelligence, University of Cambridge \\
    $^{2}$Department of Psychology, University of Cambridge \\
    $^{3}$Microsoft Research Asia
  \end{tabular}
}
\date{} 	

\renewcommand{\shorttitle}{Visuospatial Perspective Taking in MLMs}

\hypersetup{
pdftitle={Visuospatial Perspective Taking in MLMs},
pdfsubject={cs.CY, cs.AI},
pdfauthor={Jonathan Prunty, Seraphina Zhang, Patrick Quinn, Jianxun Lian, Xing Xie, Lucy Cheke},
pdfkeywords={Artificial Intelligence, Social Cognition, Theory of Mind, Spatial Cognition, Language Understanding}
}

\begin{document}

\maketitle

\begin{abstract}
As multimodal language models (MLMs) are increasingly used in social and collaborative settings, it is crucial to evaluate their perspective-taking abilities. Existing benchmarks largely rely on text-based vignettes or static scene understanding, leaving visuospatial perspective-taking (VPT) underexplored. We adapt two evaluation tasks from human studies: the Director Task, assessing VPT in a referential communication paradigm, and the Rotating Figure Task, probing perspective-taking across angular disparities. Across tasks, MLMs show pronounced deficits in Level 2 VPT, which requires inhibiting one’s own perspective to adopt another’s. These results expose critical limitations in current MLMs’ ability to represent and reason about alternative perspectives, with implications for their use in collaborative contexts.
\end{abstract}


\section{Introduction}

Generative models are becoming embedded within a wide range of social domains. They are rapidly taking on roles as teachers, colleagues, therapists, friends, and romantic partners~\citep{shevlin2024all,collins2024building}. Given this widespread adoption, it is important to prioritise the development of reliable methods for understanding the strengths and limitations of their social cognition, so as to be able to predict safe and reliable contexts of use, and avoid dangerous or costly failure cases~\citep{MIT2025,DSIT2025,bengio2025international}. 

One core feature of human social cognition is Theory of Mind (ToM): the capacity to reason about, explain, and predict others' behaviour based on their beliefs, intentions, and emotions~\citep{premack1978does, baron1985does, wimmer1983beliefs}. Given its foundational role in human social interactions~\citep{frith2007social,frith2012mechanisms}, there is growing debate about whether and to what extent such capabilities exist in artificial systems, and how they should be measured~\citep{hu2025re}. While standardised ToM tests exist within cognitive science, many require participants to read short stories and reason about the mental states of characters~\citep{happe1994advanced,baron1999recognition}. These vignette-style tasks are easily adaptable for language models, but may not reliably assess ToM capabilities in such systems. LLMs have encountered many similar story-based examples in their training corpora and can often achieve high scores by exploiting shallow associations rather than understanding underlying causal factors~\citep{mccoy2023embers,hernandez2019gazing}. Moreover, the condensed format of these stories removes much of the inferential challenge present in real social situations. Consequently, performance on vignette tasks does not always generalise to novel, naturalistic or atypical scenarios~\citep{ullman2023large,kim2023fantom,shapira2023clever}.

The advent of multimodal capabilities within language models enables evaluations using images and other sensory input, particularly paradigms from developmental and comparative psychology~\citep{frank2023baby,voudouris2025bringing}. While these often test earlier-developing ToM capabilities than adult-targeted vignette tasks, they are not inherently ``simpler'' -- indeed, they likely represent ToM challenges as experienced in real life more faithfully. In natural scenarios, our reasoning about others' minds is informed by the context in which we are situated. Relevant information is not explicitly stated in a narrative (``Elsa saw the cat'') but must be inferred from behaviour and context (e.g., perceptual input showing Elsa's eyes pointing toward a cat with no occluder between them). Visuospatial perspective taking (VPT) -- the ability to represent what someone sees and how they see it -- is among the most studied and well-understood socio-cognitive capabilities in developmental psychology. It is therefore an excellent candidate for assessing AI systems if we aim to move beyond narrow linguistic framings of Theory of Mind and generate more robust predictions about real-world AI performance. 

\subsection{Visuospatial perspective taking}

The ability to inhibit our own viewpoint and represent a scene from another's perspective is a core component of human social interactions, and is related to a range of social behaviours such as shared attention, referential communication and joint action~\citep{frith2012mechanisms, tomasello2005understanding, Clark1991grounding, sacheli2022taking}. For instance, by adopting someone's viewpoint, we can understand what they might know about their surroundings, what they might want, or what they might be referring to when speaking. Developmental psychologists have shown that this capability can be divided into two levels~\citep{flavell1981young, flavell1981development, lempers1977development, quesque2024defining} as children can determine \textit{whether} a person is seeing something (Level~1) earlier in development than they can determine \textit{how} it might appear to them (Level~2). These two capability levels can also be distinguished in adults. That is, Level~1 perspective taking judgements about whether a person can see something or not are made rapidly and spontaneously through relatively simple line-of-sight computations~\citep{samson2010seeing,o2020perspective}. By contrast, Level~2 perspective taking, involving judgements about \textit{how} something would appear to another person, is effortful and requires inhibiting one’s own egocentric perspective and mentally rotating visual information to adopt the other person's viewpoint~\citep{surtees2013similarities,surtees2016unintentional,surtees2012egocentrism,keysar2003limits,de2023perspective,michelon2006two}. Neuroimaging evidence also suggests the two levels are distinct, with Level~2 (but not Level~1) VPT tasks activating the right temporo-parietal junction (rTPJ) -- a brain region involved in distinguishing between self and other, inhibiting the egocentric perspective, and attributing mental states to others~\citep{martin2020right,perner2006thinking,schurz2014fractionating}. 

As well as making visual judgements about others' perspectives, VPT also involves making spatial judgments about \textit{where} something is located from that person’s viewpoint~\citep{surtees2013similarities,kessler2010two,quesque2024defining}. For spatial content, judging whether something is in front or behind someone is straightforward to solve by following their line of sight (Level~1), but judging whether something is to the left or right of someone is more effortful if they do not share our spatial perspective, requiring egocentric inhibition and mental rotation (Level~2). Both visual and spatial components of VPT are used in combination when interpreting referential communication~\citep{keysar2003limits,dumontheil2010taking}. For instance, when disambiguating the phrase ``Can you pass me the jar on the left?”, identifying which jar the speaker is referring to depends on whether you share the same visual perspective (as they would not be referring to a jar that only you can see) and also whether you share the same spatial perspective (if the speaker was standing opposite, their left would be your right). 

\subsection{Visuospatial perspective taking in multimodal language models}

Recent advances have enabled language models to process multiple input modalities~\citep{alayrac2022flamingo,hurst2024gpt,liu2023visual}, supporting their deployment in interactive multimodal settings such as instruction following, embodied reasoning, and collaborative task execution (see~\citet{zou2025survey} for a recent survey). Accordingly, growing attention has focused on whether multimodal language models (MLMs) can support visuospatial perspective taking (VPT).

Early investigations adapted Piaget's \textit{Three Mountains Task}~\citep{piaget1948representation} to assess Level 1 and Level 2 VPT~\citep{linsley20243d,gao2024vision}. While foundational, these benchmarks emphasise recognition of static scene layouts rather than the dynamic egocentric-to-allocentric transformations required for real-world interaction.

More recent work has introduced finer-grained paradigms. Goral and colleagues~\citep{goral2024seeing, goral2025beyond} adapted the \textit{Dot Task}~\citep{samson2010seeing}, distinguishing between visual and spatial content, but limited their evaluation to Level 1 VPT. Leonard and colleagues~\citep{leonard2024failures, leonard2025multimodal} extended this line of inquiry by adapting the \textit{Rotating Figure Task}~\citep{surtees2013similarities}, which probes both Level 1 and Level 2 VPT across viewpoint disparities ranging from 0° to 180°. They reported that MLM performance was at floor for non-shared perspectives, though their conclusions were limited by reliance on a single model (GPT-4o) and sparse trial diversity (e.g., eight trials per condition). Notably, chain-of-thought prompting improved performance only at 180°, suggesting reliance on a simple mirroring heuristic (e.g., ``swap left and right if the figure faces me'') rather than a general VPT mechanism -- a pattern deserving closer examination.

The current study systematically extends this prior work by addressing both empirical and conceptual limitations. First, we increase the rigour of the \textit{Rotating Figure Task} by evaluating a broader range of models, trials, and conditions, including several reasoning-optimised models to test whether inference-time reasoning yields genuine improvements or is reliant on simple heuristic strategies. Second, we introduce the \textit{Director Task}~\citep{keysar2003limits} to assess VPT in a functional, communicative context. This task requires models to resolve referential ambiguity based on a partner’s occluded view, providing a complementary test of how MLMs manage perspective in interactive settings.

\section{Experiments}

Our evaluation framework for visuospatial perspective taking (VPT) in multimodal language models (MLMs) adopts a procedural approach inspired by recent work creating large, controlled experimental batteries through systematic and stochastic variation of key parameters~\citep{prunty2025intuit,franken2024procedural,gandhi2023understanding}. This methodology enables comprehensive, scalable assessment while reducing benchmark contamination through novel stimulus generation and permits fine-grained control over task demands through systematic manipulation of perspective level, content type, and viewpoint disparity alongside matched control conditions.

We adapt the VIGNET approach~\citep{prunty2025intuit} to create \textit{Rotating Figure Tasks}~\citep{surtees2013similarities} -- substantially expanding prior MLM evaluations~\citep{leonard2024failures,leonard2025multimodal} -- and \textit{Director Tasks}~\citep{keysar2003limits,de2023perspective}, a referential communication paradigm not previously applied to MLMs. We test frontier MLMs from the OpenAI family, including models with (o3, o4-mini) and without (GPT-4o, gpt-4o-mini) inference-time reasoning. Full implementation details, model specifications, and supplementary analyses are provided in the appendix.

\subsection{Rotating Figure Task}

\begin{figure*}[!h]
    \centering
    \includegraphics[width=0.8\linewidth]{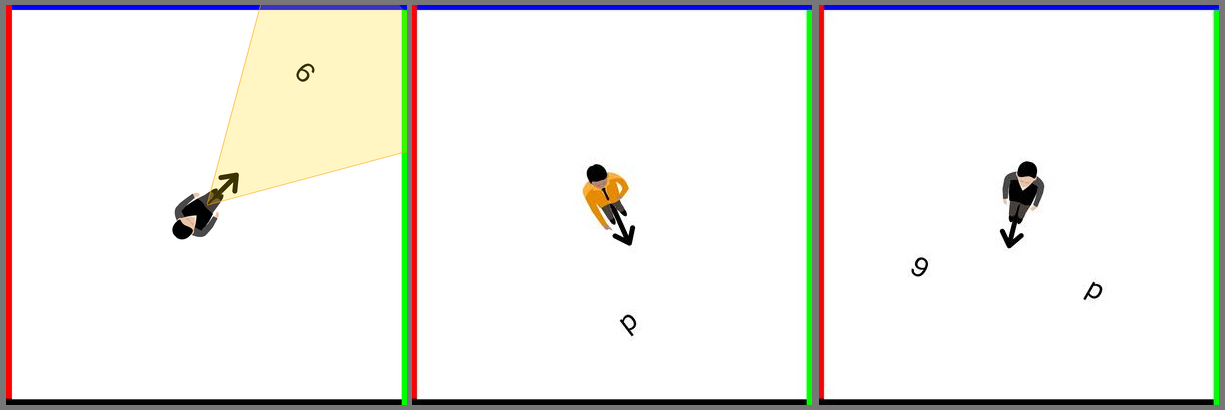}
    \caption{Example stimuli from the \textit{Rotating Figure Task} (visual field cone shown in yellow for illustration). Left: the figure \textit{can see} the symbol \textit{in front} of them (Test~1). Centre: the symbol \texttt{p} appears as a \texttt{d} from the figure’s perspective, and is located on their \textit{right} (Test~2). Right: If asked about the symbol on the figure's right appears to them, the correct response would be \texttt{6}, requiring both visual and spatial perspective taking (Test~3). A line-of-sight arrow was added to scaffold model's ability to perceive the viewing direction of figures from a top-down perspective.}
    \label{fig:vspt_example}
\end{figure*}

In the \textit{Rotating Figure Task}, participants view a top-down image of a person in a room containing a reversible symbol (e.g., \texttt{6}/\texttt{9}) placed on the floor. Questions probe either basic visual understanding (Controls 1–2) or require adopting the figure's visual or spatial perspective (Tests 1–3; Figure~\ref{fig:vspt_example}).

The critical manipulation in the \textit{Rotating Figure Task} is the angular disparity between the figure and the viewer. At \ang{0}, perspectives are fully shared; at \ang{180}, they are opposite. By varying orientation from \ang{0} to \ang{359}, we sample the full continuum of shared and non-shared perspectives. Controls 1-2 and Tests 1-3 are defined by systematically manipulating the symbol’s placement and orientation relative to the figure:

\begin{itemize}

    \item \textbf{Control~1 (Symbol identification).}  
    Identifying symbol form and location from the viewer's perspective.  
    \textit{Visual:} ``What number or letter can you see in the image?''  
    \textit{Spatial:} ``Is the number or letter on the left or right side of the image?''

    \item \textbf{Control~2 (Figure orientation).}\footnote[1]{During piloting, models struggled to infer viewing direction from top-down images. We therefore added a line-of-sight arrow (Figure~\ref{fig:vspt_example}) that scaffolds viewing-direction parsing without revealing symbol position, identity, or appearance.}  
    Identifying the direction the figure is facing.  
    \textit{Visual:} ``What colour is the wall directly in front of the person?''  
    \textit{Spatial:} ``What side of the image is directly in front of the person?''

    \item \textbf{Test~1 (Level~1 VPT: Visibility).}  
    Identifying what the figure can see: The symbol is placed in front or behind the figure. An invisible cone is used to represent their visual field (mirrored for behind-the-back placements).  
    \textit{Visual:} ``Can the person see the number or letter?''  
    \textit{Spatial:} ``Is the number or letter in front of or behind the person?''

    \item \textbf{Test~2 (Level~2 VPT: Appearance).}  
    Identifying how symbols appear from the figure's perspective and their left–right spatial relations. The symbol is positioned within the figure’s visual field but offset slightly to their left or right, and rotated to be clearly legible from the figure’s viewpoint.  
    \textit{Visual:} ``What number or letter can the person see?''  
    \textit{Spatial:} ``Is the number or letter on the person’s left or right?''

    \item \textbf{Test~3 (Level~2 VPT: Integrated visual--spatial).}  
    Joint reasoning about visual form and spatial location from the figure's perspective. Two distinct symbols are placed on the floor within the figure’s visual field, to their left and right.  
    \textit{Visuospatial:} ``What number or letter can the person see on their \{\texttt{side}\} side?''

\end{itemize}

Prior work has shown that adult humans perform this task with high accuracy, but their response times in Level 2 VPT trials increase with angular disparity \citep{surtees2013similarities}. We have extended the human paradigm by adding a more demanding test (Test~3) that requires joint visual–spatial integration under Level~2 VPT, and by introducing controls that explicitly assess precursor perceptual abilities in MLMs that are typically assumed in humans~\citep[see][]{milliere2024anthropocentric}.\footnote[2]{The final battery comprised 15k images (3k per stimulus set), yielding 27k trials due to dual visual and spatial questions in Controls 1–2 and Tests 1–2. All four models (o3, o4-mini, GPT-4o, GPT-4o-mini) completed the full battery.}

\subsubsection{Results}

Control conditions verified MLM's baseline perceptual competencies (see Appendix~\ref{appendix:supp_rotating_figure_analysis}). Models achieved high accuracy on symbol identification (Control 1), though o3 showed an elevated rate of invalid responses. Models struggled with viewing direction extraction (Control 2), particularly when figures faced corners as there is greater ambiguity as to which side of the image the figure is facing. Excluding these ambiguous corner trials substantially improved performance for most models except in GPT-4o-mini, suggesting fundamental limitations in extracting viewing direction from static images in this smaller model.

\begin{figure*}[!h]
    \hspace{1.40cm}
    \begin{center} 
    \centering
    \includegraphics[width=1\linewidth]{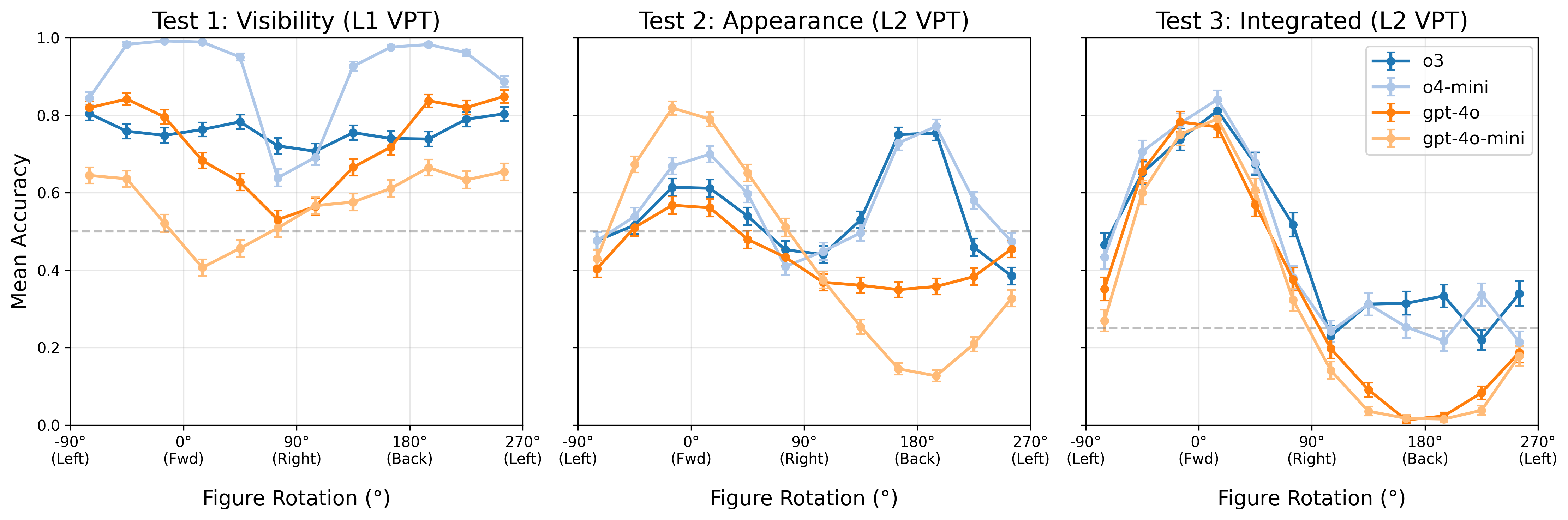}
    \caption{Mean MLM accuracy in Test conditions of the \textit{Figure Rotation Task}, binned by figure rotation angle (12 bins) and aggregated across visual and spatial questions in Tests~1–2. \ang{0} indicates a shared perspective and \ang{180} an opposite perspective; \ang{90} and \ang{-90} correspond to right- and left-facing figures. The dashed line indicates chance.}
    \label{fig:test_rotation_combined}
    \end{center}
\end{figure*}

For test conditions, Angular Disparity -- the absolute angular difference between viewer and figure perspectives -- was binned into four 45° intervals (0–45°, 45–90°, 90–135°, 135–180°). We analysed performance using mixed-effects logistic regression with Type III Wald tests and post-hoc comparisons (full models and analysis by visual and spatial question type are provided in Appendix). All main effects and interactions were significant ($p < .001$) except where noted. Figure~\ref{fig:test_rotation_combined} shows MLM's performance profiles in test conditions across the full range of rotations, while Table~\ref{tab:vpt_combined_emm} displays test performance means across increasing levels of Angular Disparity, divided by visual and spatial question type. 

\begin{table}[!h]
\centering
\caption{\label{tab:vpt_combined_emm}Angular Disparity across test conditions}
\centering
\begin{adjustbox}{max width=\columnwidth}
\begin{threeparttable}
\begin{tabular}[t]{llllllll}
\toprule
Test & Model & 0–45 & 45–90 & 90–135 & 135–180 & $\chi^2$ & p\\
\midrule
 & GPT-4o-mini & 0.40 & 0.60 & 0.68 & 0.74 & 203.6 & $<$.001\\
 & GPT-4o & 0.70 & 0.67 & 0.74 & 0.78 & 23.6 & $<$.001\\
 & o4-mini & 0.99 & 0.81 & 0.83 & 0.98 & 145.9 & $<$.001\\
\multirow[t]{-4}{*}{\raggedright\arraybackslash 1-V} & o3 & 0.77 & 0.79 & 0.77 & 0.75 & 3.5 & 0.317\\
 & GPT-4o-mini & 0.57 & 0.55 & 0.52 & 0.52 & 5.1 & 0.162\\
 & GPT-4o & 0.79 & 0.72 & 0.70 & 0.75 & 18.3 & $<$.001\\
 & o4-mini & 0.99 & 0.82 & 0.84 & 0.97 & 129.7 & $<$.001\\
\multirow[t]{-4}{*}{\raggedright\arraybackslash 1-S} & o3 & 0.74 & 0.76 & 0.75 & 0.74 & 0.8 & 0.846\\
 & GPT-4o-mini & 0.82 & 0.51 & 0.21 & 0.07 & 787.2 & $<$.001\\
 & GPT-4o & 0.31 & 0.28 & 0.34 & 0.54 & 131.2 & $<$.001\\
 & o4-mini & 0.45 & 0.28 & 0.32 & 0.57 & 160.4 & $<$.001\\
\multirow[t]{-4}{*}{\raggedright\arraybackslash 2-V} & o3 & 0.60 & 0.37 & 0.33 & 0.72 & 304.3 & $<$.001\\
 & GPT-4o-mini & 0.72 & 0.53 & 0.43 & 0.25 & 322.2 & $<$.001\\
 & GPT-4o & 0.78 & 0.59 & 0.44 & 0.19 & 493.6 & $<$.001\\
 & o4-mini & 0.86 & 0.66 & 0.63 & 0.80 & 134.7 & $<$.001\\
\multirow[t]{-4}{*}{\raggedright\arraybackslash 2-S} & o3 & 0.61 & 0.56 & 0.50 & 0.66 & 42.4 & $<$.001\\
 & GPT-4o-mini & 0.75 & 0.36 & 0.12 & 0.02 & 680.5 & $<$.001\\
 & GPT-4o & 0.75 & 0.42 & 0.17 & 0.02 & 634.7 & $<$.001\\
 & o4-mini & 0.80 & 0.47 & 0.25 & 0.28 & 493.3 & $<$.001\\
\multirow[t]{-4}{*}{\raggedright\arraybackslash 3-VS} & o3 & 0.77 & 0.52 & 0.27 & 0.31 & 420.3 & $<$.001\\
\bottomrule
\end{tabular}
\begin{tablenotes}
\item Values are estimated marginal probabilities of correct responses. $\chi^2$(3) tests the simple main effect of angular disparity within each model and task. Test labels use N–T (N = 1–3; V = visual, S = spatial, VS = visuospatial).
\end{tablenotes}
\end{threeparttable}
\end{adjustbox}
\end{table}

\textbf{Test 1 (Level 1 VPT)} revealed distinct response profiles across models. GPT-4o-mini showed monotonically increasing visual accuracy with Angular Disparity, performing below chance for aligned perspectives but above chance when inverted, while spatial performance remained near chance. GPT-4o and o4-mini exhibited U-shaped accuracy profiles, with reduced performance at intermediate disparities. For both models, this impairment was more pronounced for right-facing rotations (see Figure~\ref{fig:test_rotation_combined}). Although o4-mini approached ceiling performance for shared and inverted perspectives, it showed a marked dip for right-facing intermediate rotations. In contrast, o3 maintained stable performance across disparities and question types, indicating the greatest robustness to perspective variation.

\textbf{Test 2 (Level 2 VPT)} exposed broader limitations. GPT-4o-mini showed a systematic failure to adopt perspective, with accuracy declining sharply from aligned to inverted viewpoints (from 82\% to 7\%), as indicated by the sigmoidal performance curve in Figure~\ref{fig:test_rotation_combined}. GPT-4o exhibited qualitatively different patterns for visual versus spatial questions: spatial accuracy decreased monotonically with disparity while visual accuracy \textit{improved} once the perspective was fully inverted, suggesting inconsistent VPT strategies across modalities. Reasoning models o3 and o4-mini showed U-shaped performance across Angular Disparity with higher accuracy at aligned or inverted viewpoints but reduced accuracy at intermediate disparities. Plotting performance across the full range of rotation angles, leads to the ``M-shaped'' performance profiles in Figure~\ref{fig:test_rotation_combined}, where successful visual perspective-taking was confined to fully shared or fully opposite perspectives.

\textbf{Test 3} combined Level 2 visual and spatial perspective-taking. All models exhibited steep performance declines with increasing angular disparity (Table~\ref{tab:vpt_combined_emm}). Accuracy dropped from well above chance when perspectives aligned to near floor at maximum disparity for GPT-4o variants and approximately chance level for reasoning model. These response profiles indicate that none of the models tested reliably supports flexible, integrated visuospatial perspective-taking.

\subsection{Director Task}

\begin{figure}[!ht]
    \centering
    \includegraphics[width=0.7\linewidth]{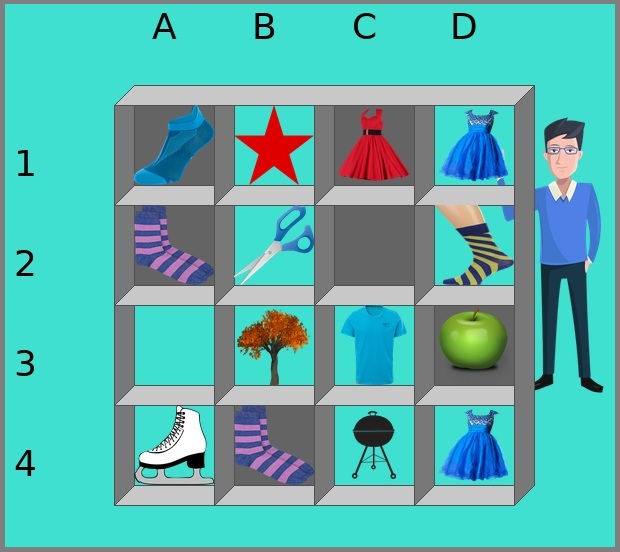}
    \caption{Example stimulus from the \textit{Director Task}. If the director says ``Please select the rightmost blue, non-striped item of clothing from my point of view'', he would be referring to the item in \texttt{C3}. The answer is not \texttt{D1} or \texttt{D4}, as the director specified the rightmost item from \textit{his} perspective. The answer is also not \texttt{A1} as the item is occluded from the director's view. All other items do not fit the object specification.}
    \label{fig:director_example}
\end{figure}

In the \textit{Director Task}~\citep{keysar2003limits}, participants follow instructions from a director positioned opposite a 4\(\times\)4 grid. Some cells have opaque backs blocking the director's view, creating visual-shared (mutually visible) and visual-different (participant-only) perspectives. In the current adaptation, participants select target objects by cell location. On visual-different trials, the director's instruction (e.g., ``select the smallest star'') requires inhibiting the viewer's privileged perspective to respond based on the director's limited access (e.g., by ignoring small stars that are occluded from the director's view). Figure~\ref{fig:director_example} provides an example trial.

For each trial, grid configurations are procedurally generated, systematically varying \textit{visual perspective} -- whether or not target items have occluded alternatives -- and \textit{relative adjective} type, which specifies target objects based on their size or spatial location (see Table~\ref{tab:rel-adj-cond}). For spatial relative adjectives, vertical terms (e.g., ``topmost'' or ``bottommost'') describe targets from a shared spatial perspective, while horizontal terms (e.g., ``leftmost'' or ``rightmost'') specify targets from inverted spatial perspectives, forming spatial-shared and spatial-different conditions, respectively.

\begin{table}[!h]
\centering
\caption{Relative adjective conditions in the Director Task}
\begin{tabularx}{1\textwidth}{llX}
\toprule
\textbf{Condition} & \textbf{Sub-type} & \textbf{Description} \\
\midrule
None    & --        & Only one object matches the director's description (e.g., ``red book'') \\
Size    & --        & Multiple objects match; target specified by size (e.g., ``the largest red book'') \\
Spatial & Shared    & Multiple matching objects; target specified by vertical term (``topmost'' or ``bottommost'') \\
Spatial & Different & Multiple matching objects; target specified by horizontal term (``leftmost'' or ``rightmost'') \\
\bottomrule
\label{tab:rel-adj-cond}
\end{tabularx}
\end{table}


The director's instructions are framed as ``from your point of view'' or ``from my point of view'', defining \textit{perspective reversal} conditions, which prevent spatial-different trials from being solved via simple heuristics (e.g., automatically reversing left–right mappings for horizontal spatial adjectives). This manipulation affects only horizontal spatial adjectives -- vertical relationships (top/bottom) remain invariant across opposing viewpoints -- for all other descriptions it is redundant. Importantly, it does not affect visual perspective: the director cannot see occluded objects and therefore would not refer to them, regardless of the ``point of view'' specified.

To examine whether limitations in basic perceptual processes such as object identification, spatial reasoning, and target localisation contribute to performance differences in this task, we developed a text-based (ASCII) version structurally identical to the image-based task (see Figure~\ref{fig:ascii-example} in Appendix). Furthermore, by comparing between versions, we can obtain convergent evidence of VPT performance across modalities.\footnote[1]{The final battery comprised 8k images (2k per dataset across four datasets with related item proportions: 0.3, 0.5, 0.7 and 0.9), yielding 16k samples across image and ASCII versions.}

\subsubsection{Results}

\begin{figure*}[!h]
    \hspace{1.40cm}
    \begin{center} 
    \centering
    \includegraphics[width=1\linewidth]{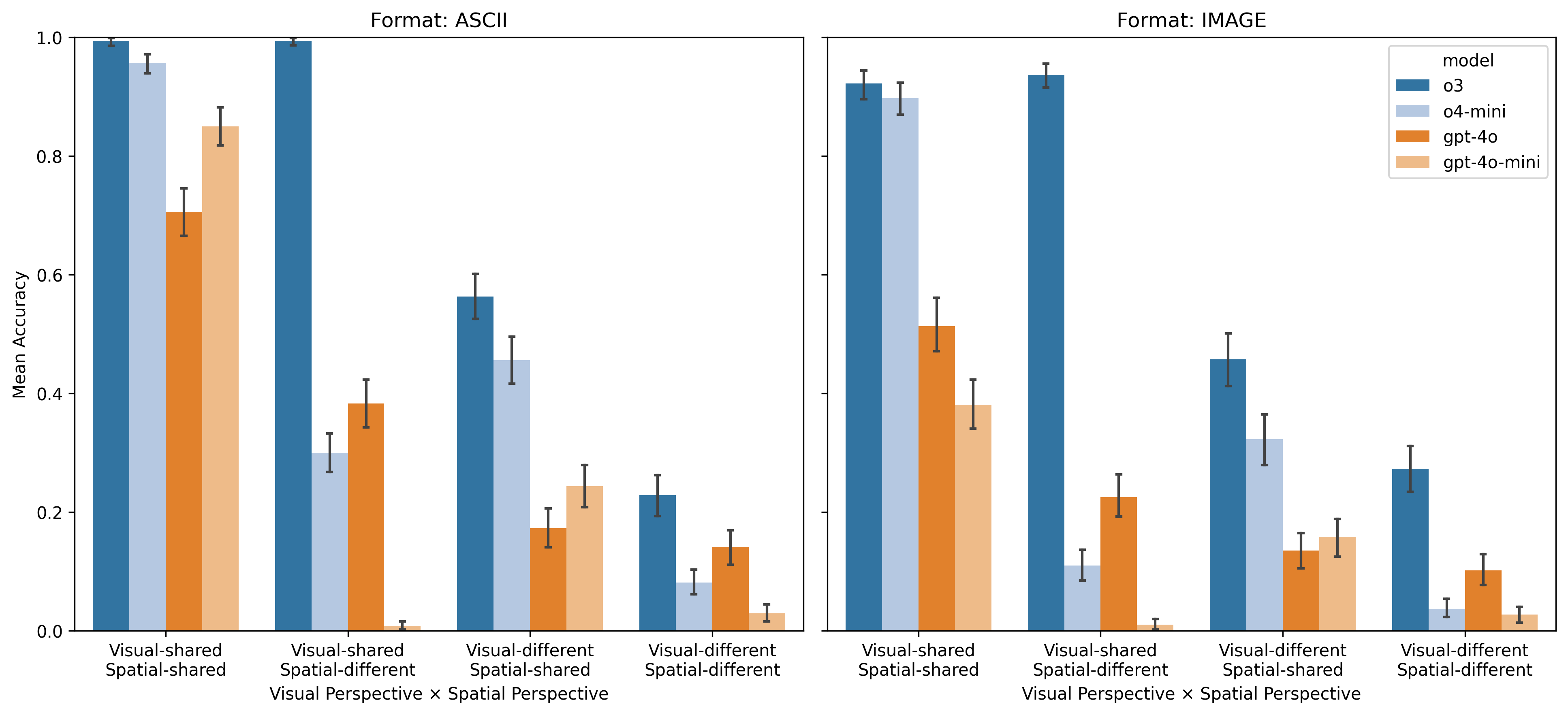}
    \caption{MLM mean accuracy in the \textit{Director Task} on visual (occluded vs. no occluded alternatives) and spatial (horizontal vs. vertical adjectives) VPT trials. The data are plotted separately for Image and ASCII tasks, and include only spatial relative adjective trials from the director's perspective. Error bars are 95 percent confidence intervals.}
    \label{fig:visualxspatial_divided}
    \end{center}
\end{figure*}

We analysed MLM performance on the \textit{Director Task} using mixed-effects logistic regression models. Our full analyses (see Appendix~\ref{appendix:supp_director_analysis}) established that accuracy was lower for image-based versus ASCII trials, indicating visual processing demands impaired performance, but similar patterns across conditions were found for both modalities (see Figure~\ref{fig:visualxspatial_divided}). Second, we found that relative adjective type (none, size, spatial-shared) did not affect baseline performance. Finally, we established that spatial PT impairments emerged specifically on horizontal adjective trials specified from the director's point of view. Given these findings, we focus here on the interaction between visual and spatial PT.   

We analysed this interaction using an AI Model $\times$ Visual Perspective $\times$ Spatial Perspective model with Task Format as a covariate, restricted to spatial relative adjective trials from the director's point of view. All main effects and interactions were significant ($p < .001$), and MLM performance is summarised in Figure~\ref{fig:visualxspatial_divided} and Table~\ref{tab:director_emm}. 

A strong main effect of AI model, $\chi^2$(3) = 529.91, revealed substantial performance differences, with reasoning models (o3, o4-mini) performing near ceiling when VPT was not required. Both Visual PT, $\chi^2$(1) = 50.91, and Spatial PT, $\chi^2$(1) = 161.97, showed significant main effects, with performance declining when perspectives differed and larger costs for spatial PT.

\begin{table}[!h]
\centering
\caption{\label{tab:director_emm}Contrasts between shared and different perspectives}
\centering
\begin{adjustbox}{max width=\columnwidth}
\begin{threeparttable}
\begin{tabular}[t]{llllllll}
\toprule
Type & Model & Shared & Different & OR & SE & p\\
\midrule
 & GPT-4o-mini & 0.10 & 0.08 & 1.37 & 0.36 & 0.242\\
 & GPT-4o & 0.45 & 0.13 & 5.30 & 1.03 & $<$.001\\
 & o4-mini & 0.66 & 0.17 & 9.46 & 1.91 & $<$.001\\
\multirow[t]{-4}{*}{\raggedright\arraybackslash Visual} & o3 & 0.97 & 0.37 & 52.87 & 11.39 & $<$.001\\
 & GPT-4o-mini & 0.39 & 0.01 & 43.62 & 11.60 & $<$.001\\
 & GPT-4o & 0.35 & 0.19 & 2.27 & 0.44 & $<$.001\\
 & o4-mini & 0.76 & 0.11 & 25.04 & 5.05 & $<$.001\\
\multirow[t]{-4}{*}{\raggedright\arraybackslash Spatial} & o3 & 0.85 & 0.77 & 1.69 & 0.36 & 0.015\\
\bottomrule
\end{tabular}
\begin{tablenotes}
\item Shared/Different are estimated marginal probabilities. OR = odds ratio, SE = standard error and p from Shared/Different pairwise contrasts.
\end{tablenotes}
\end{threeparttable}
\end{adjustbox}
\end{table}

Crucially, the magnitude of visual and spatial PT effects varied across models, as shown by significant AI Model × Visual PT, $\chi^2$(3) = 85.18, AI Model × Spatial PT, $\chi^2$(3) = 449.09, and three-way, $\chi^2$(3) = 134.52, interactions. o3 showed a pronounced accuracy drop under visual PT demands but remained comparatively robust on spatial PT trials. In contrast, other models were substantially impaired under both forms of VPT: o4-mini and GPT-4o-mini were more affected by spatial PT, whereas GPT-4o showed larger decrements under visual PT. Performance declined further when both demands were combined, most clearly for o3, as other models were already near floor under single-demand conditions. Overall, these results indicate persistent VPT limitations despite strong baseline performance, with model-specific patterns suggesting distinct processing strategies or constraints.

\section{Discussion}

Visuospatial perspective-taking (VPT) is central to social interaction. We evaluated VPT in frontier multimodal language models using two complementary paradigms: the \textit{Rotating Figure Task}, which isolated perspective-taking across angular disparity, content (visual vs. spatial), and type (Level 1 vs. Level 2), and the \textit{Director Task}, which assessed visual and spatial VPT within a referential communication paradigm. Across both tasks, all models exhibited persistent VPT limitations, with distinct failure patterns that reveal the mechanisms underlying their performance.

Despite strong performance on perceptual controls, systematic VPT deficits emerged. In the \textit{Rotating Figure Task}, even Level 1 VPT (visibility judgements) revealed model-specific failures, including right-facing spatial blind spots (GPT-4o, o4-mini) and frequent invalid responses due to hallucinations (o3). Level 2 VPT exposed more severe impairments. GPT models showed largely monotonic accuracy declines with increasing angular disparity, whereas reasoning models exhibited M-shaped performance profiles: success at 0° and 180°, but failures at intermediate rotations. When visual and spatial perspective-taking were required together, all models fell to chance or floor performance as perspectives diverged.

The \textit{Director Task} provided convergent evidence in an applied communicative setting. Both visual perspective-taking (occlusion tracking) and spatial perspective-taking (left–right reversals) reduced performance, with larger costs for spatial transformations. Similar patterns across image-based and ASCII versions indicate that these failures reflect cross-modal VPT limitations rather than specifically visual processing deficits. Notably, only o3's (not o4-mini's) preserved performance for fully-opposite (i.e., \ang{180}) spatial perspectives in the \textit{Rotating Figure Task} translated into robust spatial VPT on the \textit{Director Task}, highlighting differences in how models generalize VPT to functional contexts.

Together, these patterns suggest reliance on shallow heuristics. The M-shaped accuracy profiles observed for reasoning models in Level 2 VPT likely reflect simple mirroring strategies: swapping left and right at 180°, or applying symbol inversion (e.g., 6 becomes 9), but lack the representational capacity for intermediate transformations requiring genuine mental rotation. While extended inference-time reasoning or chain-of-thought prompting can improve the application of such heuristics, the underlying representational limitations remain. Such strategies fail in applied contexts like the \textit{Director Task}, where VPT is employed in service of another goal -- identifying and selecting the correct referent while simultaneously tracking occlusions and disambiguating spatial references. Here, all models approached floor performance when visual and spatial VPT demands were combined.

\subsection{Related work}

Our findings substantially expand prior work investigating VPT in MLMs. \citet{leonard2024failures, leonard2025multimodal} found that GPT-4o showed improved performance at 180° disparity with chain-of-thought prompting. We demonstrate systematically that inference-time reasoning models show qualitatively different performance profiles than non-reasoning models, but fundamental limitations remain -- particularly when multiple VPT demands combine. \citet{goral2024seeing, goral2025beyond} showed impaired Level 1 VPT in GPT-4o and older models, with particularly poor spatial VPT. Our findings for GPT-4o are consistent but also reveal underlying mechanisms: blind spots for right-facing rotations, heuristic-like performance spikes at 180° for visual VPT, but monotonic declines for spatial VPT. Critically, we also demonstrate that these VPT deficits translate to failures in disambiguating referential communication in applied settings.

Comparisons with human VPT further highlight these differences. In humans, Level 2 judgements are effortful, scale with angular disparity, and recruit distinct neural mechanisms including rTPJ to inhibit egocentric representations \citep{surtees2013similarities,martin2020right}, whereas Level 1 visibility judgements are comparatively automatic \citep{samson2010seeing}. Models, like humans, showed a similar distinction for Level 2 relative to Level 1 VPT, though their error patterns were often non-human-like, including spatial blind spots, divergent performance for visual versus spatial content, and disproportionate failures at intermediate rather than maximal disparities. These tasks therefore offer a basis for future comparative work across human and artificial systems, including mechanistic analyses to identify architectural components supporting Level 1 versus Level 2 VPT, potentially paralleling dissociable neural substrates in humans.

\subsection{Conclusion}

These persistent VPT limitations have practical implications for deploying MLMs in social and collaborative settings. Perspective-taking supports referential communication, spatial coordination, and joint action \citep{Clark1991grounding, frith2012mechanisms}. Models that cannot flexibly represent what others can see will struggle with collaborative task execution, instruction-following in shared environments, and embodied reasoning \citep{zou2025survey}. Moreover, the failure modes identified here may be hidden in casual interaction or simple benchmark scores, but emerge under particular geometric configurations or combined VPT demands.

Overall, we show that evaluating multimodal models using controlled visuospatial paradigms provides a  precise characterization of their socio-cognitive capacities and clearer guidance for deployment. As MLMs become embedded in human social and physical environments, rigorous assessment of foundational abilities like visuospatial perspective-taking will be essential for safe and reliable use.

\subsubsection*{Acknowledgments}
The authors acknowledge the support of Accenture  (https://www.accenture.com/us-en/services/data-ai) and Microsoft Research Asia (https://www.microsoft.com/en-us/research/lab/microsoft-research-asia/) to this research. This research project has also benefitted from the Microsoft Accelerate Foundation Models Research (AFMR) grant program.

\bibliographystyle{unsrtnat}
\bibliography{main} 

\newpage
\appendix

\section{Task Details}
\label{appendix:task_details}

\newtcolorbox{msgbox}[3][]{%
  enhanced, breakable,
  arc=3pt, boxrule=0.6pt,
  colframe=#2!80!black, colback=white,
  left=6pt, right=6pt, top=6pt, bottom=6pt,
  coltitle=white, fonttitle=\bfseries,
  colbacktitle=#2,
  title={#3},
  #1
}
\newcommand{\user}[1]{\begin{msgbox}{blue}{User}#1\end{msgbox}}
\newcommand{\assistant}[1]{\begin{msgbox}{green!60!black}{Assistant}#1\end{msgbox}}
\newcommand{\system}[1]{\begin{msgbox}{orange}{System}#1\end{msgbox}}

In this section, we will provide additional details of the tasks, prompts and tooling available to the models tested in our experiments. Each of the tasks were created using the Inspect AI framework \cite{UK_AI_Security_Institute_Inspect_AI_Framework_2024} to create a ReAct \citep{yao2022react} loop where an AI agent can call a think() tool to perform chain-of-thought reasoning before calling a submit() tool to provide their answer. The generic system prompts used are provided in Table~\ref{tab:prompts}.

\begin{table}[!h]
\centering
\caption{System prompts}
\label{tab:prompts}
\small
\begin{tabular}{>{\bfseries}lp{10cm}}
\toprule
Prompt & Content \\
\midrule
General &
  {\ttfamily You are a helpful assistant attempting to submit the correct answer. You have several functions available to help with finding the answer. Each message may perform one function call. You will see the result of the function right after sending the message. If you need to perform multiple actions, you can always send more messages with subsequent function calls. Do some reasoning before your actions, describing what function calls you are going to use and how they fit into your plan. When you have completed the task and have an answer, call the submit() function to report it.} \\
\midrule
Rotating Figure Task &
  {\ttfamily You are participating in a visual perspective-taking experiment. You will be shown images containing a person (or arrow) and possibly a number or symbol on the ground nearby. Your task is to answer questions about what you can see in the image, considering both your own perspective as the viewer and the perspective of the person/figure in the image. Pay careful attention to: the orientation and position of figures in the image; whether numbers appear as ``6'' or ``9'' depending on viewing angle; spatial relationships (left/right, front/behind) from different perspectives; what is visible vs.\ hidden from different viewpoints. Answer exactly as requested in the prompt.} \\
\midrule
Director Task &
  {\ttfamily There is a shelf in front of you with a grid of items on it. The user is standing on the opposite side of the shelves and is asking you to pick up a specific item for them that they can see. Some of the cells in the grid are blocked from the user's view, hiding anything that might be there, indicated by the dark background in the cell. You will receive an image of the grid from your point of view. Submit your answer in the form \textless column\textgreater\textless row\textgreater, for example `C2'.} \\
\bottomrule
\end{tabular}
\end{table}

The full framework for generating stimuli for both the Rotating Figure Task and the Director Task, for running the experiments and analysing the results are provided in the project repo\footnote{https://github.com/JonnyP1990/visual-perspective-taking}. The generated stimuli used within our experiments are also available \href{https://osf.io/bpr5j/overview?view_only=d583d77e39394af08f07ae69881bb938}{here}\footnote{https://osf.io/bpr5j/overview?view\_only=d583d77e39394af08f07ae69881bb938} and each is password protected. The password for the directors task dataset is \verb|director_task| and the one for the rotating figure task is \verb|vspt_task|.

\begin{SaveVerbatim}{SubmitFunc}
{
  "name": "submit",
  "description": "Submit an answer for evaluation.",
  "parameters": {
    "type": "object",
    "properties": {
      "answer": {
        "type": "string",
        "description": "Submitted answer"
      }
    },
    "required": ["answer"],
    "additionalProperties": false
  }
}
\end{SaveVerbatim}

\begin{SaveVerbatim}{ThinkFunc}
{
  "name": "think",
  "description": "Use this tool to stop, think and reason about
  the task at hand. Particularly useful to build plans and reflect
  on them prior to executing actions.",
  "parameters": {
    "type": "object",
    "properties": {
      "thoughts": {
        "type": "string",
        "description": "The thoughts or reasoning process of the agent."
      }
    },
    "required": ["thoughts"],
    "additionalProperties": false
  }
}
\end{SaveVerbatim}

\begin{table}[!h]
\centering
\caption{Description of stimulus sets for the Rotating Figure Task}
\label{tab:rft-stimulus-descriptions}
\small
\begin{tabular}{>{\bfseries}lp{11cm}}
\toprule
Stimulus Set & Description \\
\midrule
Control 1 &
  Assessed symbol identification and left--right spatial discrimination. Non-rotated symbols were presented either to the left or right of the figure. Figure orientation was constrained to 70°--110° to ensure that symbols were positioned unambiguously on the left or right side of the image. \\
\midrule
Control 2 &
  Assessed line-of-sight judgments. Participants were required to identify either the colour or the spatial location of the wall directly in front of the figure. Figure orientation varied across the full 360° range. Symbols were masked in this condition. \\
\midrule
Test 1 &
  Assessed Level 1 visual perspective taking (VPT), specifically whether a symbol was visible to the figure. Symbols were positioned either in front of or behind the figure, defined relative to a 60° field-of-view (FOV) cone extending from the figure. Symbols were oriented to appear either upright or inverted from the figure's viewpoint, with a small amount of rotational jitter applied. Figure orientation varied across the full 360°. \\
\midrule
Test 2 &
  Assessed Level 2 VPT, requiring judgments about how a symbol appeared or where it was located from the figure's perspective. Trial structure was identical to Test 1, except that symbol placement was restricted to either the front-left or front-right of the FOV. \\
\midrule
Test 3 &
  Assessed combined visual and spatial Level 2 VPT. Trial structure was identical to Test 2, except that two symbols were presented simultaneously: one in the front-left region and one in the front-right region of the FOV cone. The second symbol always differed in identity from the first (e.g., 6 vs m). \\
\bottomrule
\end{tabular}
\end{table}

\begin{table}[!h]
\centering
\caption{Stimulus set prompts for the Rotating Figure Task}
\label{tab:rft-stimulus-prompts}
\small
\begin{tabular}{>{\bfseries}lp{3.5cm}p{4cm}p{4cm}}
\toprule
Stimulus Set & Context Prompt & Visual Prompt & Spatial Prompt \\
\midrule
control\_1 &
  {\ttfamily There is a person standing in the centre of the image, and a number or letter on the floor nearby.} &
  {\ttfamily What number or letter can you see in the image? Respond with a single number or letter:} &
  {\ttfamily Is the number or letter on the left or right side of the image? Respond with a single word: LEFT or RIGHT} \\
\midrule
control\_2 &
  {\ttfamily There is a person standing in the centre of the image. There are coloured walls along the border.} &
  {\ttfamily What colour is the wall directly in front of the person? Respond with a single word: RED, GREEN, BLUE or BLACK} &
  {\ttfamily Which side of the image is directly in front of the person? Respond with a single word: LEFT, RIGHT, TOP or BOTTOM} \\
\midrule
test\_1 &
  {\ttfamily There is a person standing in the centre of the image, and a number or letter on the floor nearby.} &
  {\ttfamily Can the person see the number or letter? Respond with either: CAN SEE or CANNOT SEE} &
  {\ttfamily Is the number or letter in front of or behind the person? Respond with a single word: FRONT or BEHIND} \\
\midrule
test\_2 &
  {\ttfamily There is a person standing in the centre of the image, and a number or letter on the floor nearby.} &
  {\ttfamily What number or letter can the person see? Respond with a single number or letter:} &
  {\ttfamily Is the number or letter on the person's left or right? Respond with a single word: LEFT or RIGHT} \\
\midrule
test\_3 &
{\ttfamily There is a person standing in the centre of the image, and 2 numbers or letters on the floor nearby.} &
\multicolumn{2}{p{8cm}}{%
\ttfamily What number or letter can the person see on their \{side\} side?
Respond with a single number or letter:} \\
\bottomrule
\end{tabular}
\end{table}

The Rotating Figure Task (RFT) comprised two control conditions and three test conditions. The project repository contains the scripts and parameter settings used to generate 3,000 items per condition and visual/spatial combination (27,000 items in total; Test 3 included only a single visuospatial prompt). The prompts used in this task are listed in Table~\ref{tab:rft-stimulus-prompts}, and detailed descriptions of each stimulus set are provided in Table~\ref{tab:rft-stimulus-descriptions}.

In the Director Task (DT), prompts were presented as user messages requesting that the model select a target item from a grid. Targets could be distinguished from matched distractors based either on visual perspective (i.e., whether the item was occluded from the “user”) or on a relative adjective in the description, which was either spatial (e.g., “leftmost”) or size-based (e.g., “largest”). In addition to matched distractors, grids also contained non-matching distractors, which could be excluded because they did not satisfy the target description.

The object images and their associated properties used to generate item descriptions are available in the project repository (see \texttt{items.json}). Properties included both physics-related attributes (e.g., “holds water”) and non-physics attributes (e.g., “clothing”). During stimulus generation, fill proportion determined the proportion of grid cells containing an item, and related item proportion determined the proportion of distractors matching the target description. Item difficulty was manipulated by generating denser grids (higher fill proportion) with a greater number of matched distractors (higher related item proportion). The final battery comprised 8,000 images: 2,000 per dataset across four datasets with related item proportions of 0.3, 0.5, 0.7, and 0.9. Across both image-based and ASCII versions, this yielded 16,000 total samples.

To complement the image-based Director Task, we generated a structurally equivalent text-based (ASCII) version. The same underlying dataset was used for both the ASCII and image-based formats. Figure~\ref{fig:ascii-example} displays the human-readable rendering of the ASCII grid for illustrative purposes; however, models received the raw, unrendered markdown representation of this grid. The following sections present example transcripts for both task formats.

\begin{figure*}[h!]
  \begin{center}
  \begin{verbatim}
    ================================================================================                        
  | A1                 | B1                 | C1                 | D1                 |                   
  | [BLOCKED]          | blue_book_small    | blue_book_small    | blue_book_small    |                   
  | cookie             | B: blue,stackable, | B: blue,stackable, | B: blue,stackable, |                   
  |                    |    book            |    book            |    book            |                   
  | S: size:1          | S: size:0          | S: size:0          | S: size:0          |                   
  -------------------------------------------------------------------------------------                   
  | A2                 | B2                 | C2                 | D2                 |                   
  | [BLOCKED]          | [BLOCKED]          | blue_book_small    | blue_book_small    |                   
  | blue_book_small    | blue_book_small    |                    |                    |                   
  -------------------------------------------------------------------------------------                   
  | A3                 | B3                 | C3                 | D3                 |                   
  | beatbox            | kitchen_knife      | blue_book          | blue_book_small    |                   
  | S: size:1          | S: size:1          | S: size:1          | S: size:0          |                   
  -------------------------------------------------------------------------------------                   
  | A4                 | B4                 | C4                 | D4                 |                   
  | [BLOCKED]          | [BLOCKED]          | [BLOCKED]          | Empty              |                   
  | Empty              | black_pot_small    | blue_book_small    |                    |                   
  ================================================================================ 
\end{verbatim}
\end{center}
\caption{Example grid from the ASCII version of the Director Task, shown as rendered from the underlying markdown representation. Item descriptors and attributes (e.g., size) are displayed explicitly, and occluded items are marked with \texttt{[BLOCKED]}.}
\label{fig:ascii-example}
\end{figure*}



\clearpage
\subsection{Rotating Figure Task transcript}
\system{You are participating in a visual perspective-taking experiment. You will be shown images containing a person (or arrow) and possibly a number or symbol on the ground nearby.

Your task is to answer questions about what you can see in the image, considering both your own perspective as the viewer and the perspective of the person/figure in the image.

Pay careful attention to:
\begin{itemize}
    \item The orientation and position of figures in the image
    \item Whether numbers appear as "6" or "9" depending on viewing angle
    \item Spatial relationships (left/right, front/behind) from different perspectives
    \item What is visible vs. hidden from different viewpoints
\end{itemize}
Answer exactly as requested in the prompt.}
\system{You are a helpful assistant attempting to submit the correct answer. You have
several functions available to help with finding the answer. Each message
may perform one function call. You will see the result of the function right
after sending the message. If you need to perform multiple actions, you can
always send more messages with subsequent function calls. Do some reasoning
before your actions, describing what function calls you are going to use and
how they fit into your plan.

When you have completed the task and have an answer, call the submit()
function to report it.}
\user{There is a person standing in the centre of the image, and a number or letter on the floor nearby.
What number or letter can you see in the image?

Respond with a single number or letter:}
\user{\includegraphics[width=0.4\linewidth]{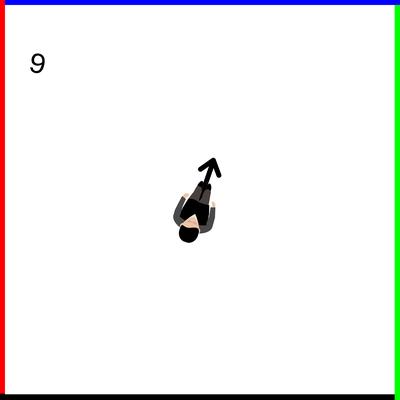}}
\assistant{submit(answer : "9")}

\clearpage
\subsubsection{Director Task transcript}

\system{The image displays the front of a shelving grid, which has items stored in grid cells. Your task is to select the correct grid cell that contains the item specified by the user. Each grid cell has a backing plate behind the item that is either transparent or made from an opaque grey material. The user is standing on the opposite side of the grid, viewing it from the back. The user has never seen the grid before, and has no prior knowledge as to the location of the items.

Submit your answer in the form <column><row>, for example 'C2' where A is the leftmost column and 1 is the topmost row. Give the coordinates from your point of view.}
\system{You are a helpful assistant attempting to submit the correct answer. You have
several functions available to help with finding the answer. Each message
may perform one function call. You will see the result of the function right
after sending the message. If you need to perform multiple actions, you can
always send more messages with subsequent function calls. Do some reasoning
before your actions, describing what function calls you are going to use and
how they fit into your plan.

When you have completed the task and have an answer, call the submit()
function to report it.}
\user{Please select the largest star from your point of view}
\user{\includegraphics[width=0.6\linewidth]{Images/fig_2.jpg}}
\assistant{submit(answer : "B1")}

\clearpage
\section{Supplementary analyses}
\label{appendix:supplementary_analyses}

\subsection{Rotating Figure Task}
\label{appendix:supp_rotating_figure_analysis}

\subsubsection{Controls 1 and 2}
\begin{figure}[ht]
    \hspace{1.40cm}
    \begin{center} 
    \centering
    \includegraphics[width=0.7\linewidth]{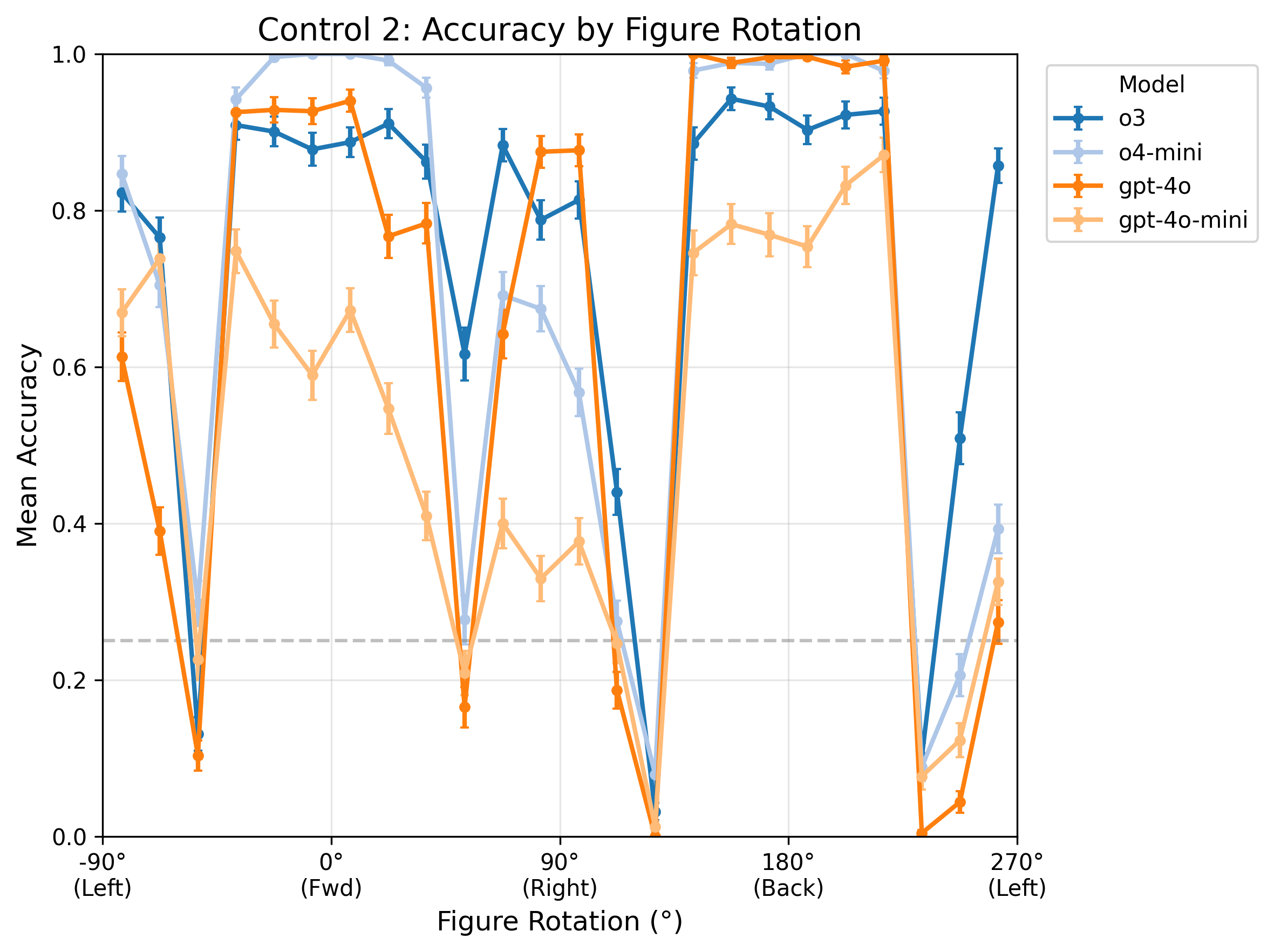}
    \caption{Mean MLM accuracy in Control 2 of the \textit{Figure Rotation Task}. Accuracy is aggregated into 24 bins by figure rotation angle. At \ang{0}, the figure faces forward (shared perspective); at \ang{180}, it faces backward (opposite perspective). Angles wrap around such that \ang{90} corresponds to the figure facing right and \ang{-90} (=\ang{270}) corresponds to the figure facing left. The dashed line represents chance performance.}
    \label{fig:control_rotation}
    \end{center}
\end{figure}

Before evaluating performance on the test conditions of the \textit{Rotating Figure Task}, we verified that MLMs exhibited the baseline capabilities required by the task. Specifically, Control 1 assessed symbol identity recognition and spatial localisation (e.g., \texttt{6}/\texttt{9}, \texttt{Left}/\texttt{Right}), while Control 2 assessed basic line-of-sight judgements (e.g., figure is facing the \texttt{Left}/\texttt{Red} wall). Table~\ref{tab:control_accuracy} and Figure~\ref{fig:control_rotation} summarise the mean accuracy scores across both control tasks, divided by question type (visual and spatial). 

\begin{table}[!h]
\centering
\caption{\label{tab:control_accuracy}Mean accuracy in control conditions}
\centering
\resizebox{\ifdim\width>\linewidth\linewidth\else\width\fi}{!}{
\begin{tabular}[t]{lcccccccc}
\toprule
 & \multicolumn{4}{c}{Control 1} & \multicolumn{4}{c}{Control 2} \\
 & \multicolumn{2}{c}{Accuracy} & \multicolumn{2}{c}{Invalid} & \multicolumn{2}{c}{Accuracy} & \multicolumn{2}{c}{Accuracy (CR)} \\
Model & Visual & Spatial & Visual & Spatial & Visual & Spatial & Visual & Spatial \\
\midrule
o3 & 0.824 & 0.853 & 0.048 & 0.106 & 0.716 & 0.752 & 0.801 & 0.843 \\
o4-mini & 0.886 & 0.978 & 0.037 & 0.002 & 0.698 & 0.715 & 0.774 & 0.798 \\
gpt-4o & 0.912 & 1.000 & 0.048 & 0.000 & 0.594 & 0.696 & 0.663 & 0.787 \\
gpt-4o-mini & 0.901 & 0.979 & 0.041 & 0.000 & 0.403 & 0.608 & 0.435 & 0.679 \\
\bottomrule
\multicolumn{9}{l}{\textsuperscript{} CR = corners removed (trials within \ang{10} of a corner [21.8\%] excluded)}\\
\multicolumn{9}{l}{\textsuperscript{} Invalid = proportion of responses not matching any valid answer option}\\
\end{tabular}}
\end{table}

MLMs showed high accuracy on Control 1 overall, with gpt-4o, for example, achieving perfect performance on spatial questions. Performance on spatial questions was slightly lower for o3 relative to other models ($M$ = 0.85; all other MLMs $M$ > 0.97). This difference is partly attributable to o3’s elevated invalid response rate ($M$ = 0.11; all other MLMs $M$ < 0.01), with o3 frequently producing verbose or irrelevant outputs despite prompts that explicitly constrained responses (e.g., \texttt{Left} or \texttt{Right}).

Despite the inclusion of an arrow indicating viewing direction in the stimuli (Figure~\ref{fig:vspt_example}), MLMs showed difficulty in reliably parsing the figure’s viewing direction in Control 2. Because this ability was evaluated across the full range of rotation angles, trials in which the figure faced toward image corners (\ang{-45}, \ang{45}, \ang{135}, \ang{225}; Figure~\ref{fig:control_rotation}) introduced inherent ambiguity regarding which side of the image the figure was facing. Consistent with this, performance exhibited pronounced decreases at these rotation angles. When ambiguous trials within \ang{10} of a corner (21.8\% of trials) were excluded, mean accuracy increased substantially for all models except gpt-4o-mini. This pattern suggests that, unlike the other models, gpt-4o-mini may have more fundamental limitations in extracting viewing direction from static images.

\subsubsection{Test 1: Level 1 VPT}
For the test conditions of the \textit{Rotating Figure Task}, MLM performance was analysed using mixed-effects logistic regression models. Type III Wald $\chi^2$ tests were used to assess main and interaction effects, followed by post-hoc comparisons of estimated marginal means to examine simple main effects across Angular Disparity bins. Model summaries are reported in Tables~\ref{tab:vpt_model1_summary},~\ref{tab:vpt_model2_summary}, and~\ref{tab:vpt_model3_summary} for Test conditions 1, 2, and 3, respectively.

\begin{figure*}[ht]
    \hspace{1.40cm}
    \begin{center} 
    \centering
    \includegraphics[width=1\linewidth]{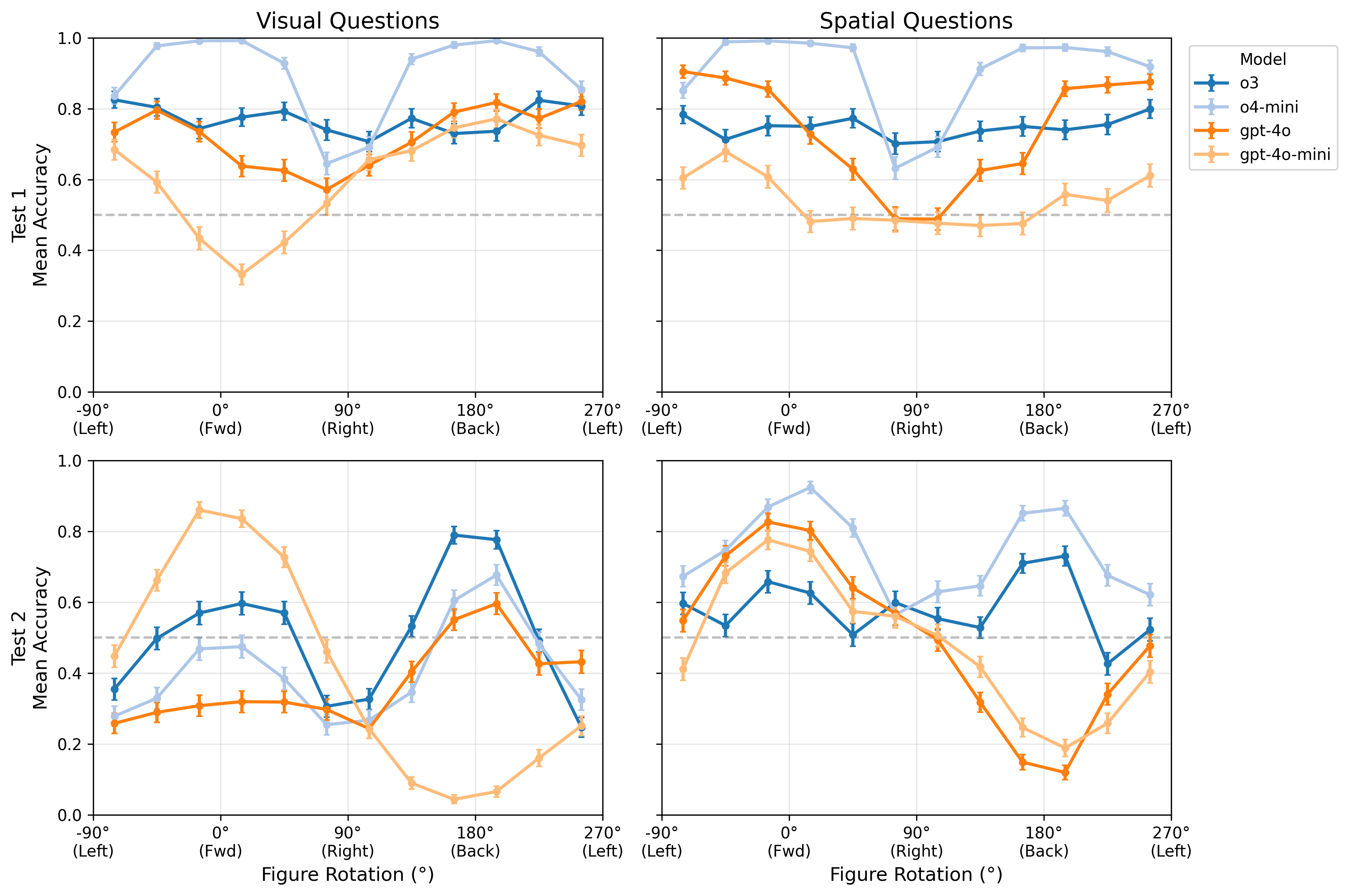}
    \caption{Mean MLM accuracy in Test 1 and 2 of the \textit{Figure Rotation Task}, by Question Type (Visual vs. Spatial). Accuracy is aggregated into 12 bins by figure rotation angle. At \ang{0}, the figure faces forward (shared perspective); at \ang{180}, it faces backward (opposite perspective). Angles wrap around such that \ang{90} corresponds to the figure facing right and \ang{-90} (=\ang{270}) corresponds to the figure facing left. The dashed line represents chance performance.}
    \label{fig:test_rotation}
    \end{center}
\end{figure*}

Angular Disparity served as the critical manipulation of perspective. It was defined as the absolute angular difference between the participant’s viewing perspective and the figure’s facing direction, constrained to a 0–180° range. Rotations exceeding 180° were subtracted from 360° to yield a symmetric measure of deviation from the forward-facing orientation (0° = facing toward the viewer; 180° = facing away), irrespective of rotational direction. For modelling, Angular Disparity was binned into four 45° intervals (0–45°, 45–90°, 90–135°, and 135–180°), represented by their midpoint values (22.5°, 67.5°, 112.5°, and 157.5°).

In Test 1, symbols appeared either in front of or behind the figure (Figure~\ref{fig:vspt_example}). Visual questions asked whether the figure could see the symbol, whereas spatial questions asked whether the symbol was in front of or behind the figure. Figure~\ref{fig:test_rotation} shows model accuracy as a function of figure rotation for both question types. Table~\ref{tab:vpt_model1_emm} reports estimated marginal means for Angular Disparity bins and the corresponding simple main effects tests.

\begin{table}[!h]
\centering
\caption{\label{tab:vpt_model1_emm}Mean accuracy in Test 1 by Angular Disparity}
\centering
\resizebox{\ifdim\width>\linewidth\linewidth\else\width\fi}{!}{
\begin{tabular}[t]{lllllllll}
\toprule
Question & Model & 0–45° & 45–90° & 90–135° & 135–180° & $\chi^2$ & df & p\\
\midrule
 & GPT-4o-mini & 0.401 & 0.601 & 0.682 & 0.745 & 203.59 & 3 & $<$ .001\\

 & GPT-4o & 0.699 & 0.672 & 0.735 & 0.779 & 23.57 & 3 & $<$ .001\\

 & o4-mini & 0.986 & 0.812 & 0.830 & 0.978 & 145.93 & 3 & $<$ .001\\

\multirow[t]{-4}{*}{\raggedright\arraybackslash Visual} & o3 & 0.771 & 0.792 & 0.771 & 0.751 & 3.53 & 3 & 0.317\\

 & GPT-4o-mini & 0.569 & 0.549 & 0.517 & 0.525 & 5.14 & 3 & 0.162\\

 & GPT-4o & 0.788 & 0.720 & 0.697 & 0.750 & 18.26 & 3 & $<$ .001\\

 & o4-mini & 0.991 & 0.824 & 0.836 & 0.973 & 129.74 & 3 & $<$ .001\\

\multirow[t]{-4}{*}{\raggedright\arraybackslash Spatial} & o3 & 0.737 & 0.755 & 0.751 & 0.743 & 0.81 & 3 & 0.846\\
\bottomrule
\multicolumn{9}{l}{\textsuperscript{} Values are estimated marginal probabilities of correct responses.}\\
\multicolumn{9}{l}{\textsuperscript{} Simple main effects test angular disparity within each Model $\times$ Question Type combination.}\\
\end{tabular}}
\end{table}

We observed significant main effects of Model, Question Type, and Angular Disparity, as well as significant interactions among these factors (all $p < .001$; Table~\ref{tab:vpt_model1_summary}), indicating that the effect of Angular Disparity differed across models and question types. For GPT-4o-mini, visual question accuracy increased monotonically with Angular Disparity, rising from below chance when perspectives were aligned (0–45°: $M = .40$) to substantially higher accuracy when perspectives were inverted (135–180°: $M = .75$; $\chi^2(3) = 203.59$, $p < .001$). In contrast, spatial question performance for GPT-4o-mini remained near chance across all disparity bins (range $M = .52$–.57), with no reliable effect of Angular Disparity ($\chi^2(3) = 5.14$, $p = .162$).

GPT-4o and o4-mini both exhibited U-shaped performance profiles, characterised by reduced accuracy at intermediate disparities (45–135°). For GPT-4o, visual accuracy was lowest when the figure faced left or right (45–90°: $M = .67$; 90–135°: $M = .74$) and higher when perspectives were aligned or inverted (0–45°: $M = .70$; 135–180°: $M = .78$; $\chi^2(3) = 23.57$, $p < .001$). A similar but more pronounced pattern was observed for o4-mini, which approached ceiling performance at low and high disparity for both visual (0–45°: $M = .99$; 135–180°: $M = .98$) and spatial questions (0–45°: $M = .99$; 135–180°: $M = .97$), with marked reductions at intermediate angles (visual: $M \approx .81$–.83; spatial: $M \approx .82$–.84; both $p < .001$).

In contrast, o3 showed the greatest robustness to both content and perspective variation. For this model, accuracy remained relatively stable across Angular Disparity bins for both visual (range $M = .75$–.79; $\chi^2(3) = 3.53$, $p = .317$) and spatial questions (range $M = .74$–.76; $\chi^2(3) = 0.81$, $p = .846$), indicating no reliable effect of figure rotation.

\subsubsection{Test 2: Level 2 VPT}

In Test 2, symbols were positioned in front of the figure and deviated slightly to the left or right, while remaining within the figure’s field of view. Visual questions probed how the symbol appeared from the figure’s perspective (e.g., whether it appeared as a 6 or a 9), whereas spatial questions asked whether the symbol was to the figure’s left or right. Figure~\ref{fig:test_rotation} shows accuracy as a function of figure rotation for both question types, and Table~\ref{tab:vpt_model2_emm} reports estimated marginal means across Angular Disparity bins along with simple main effects tests.

\begin{table}[!h]
\centering
\caption{\label{tab:vpt_model2_emm}Mean accuracy in Test 2 by Angular Disparity}
\centering
\resizebox{\ifdim\width>\linewidth\linewidth\else\width\fi}{!}{
\begin{tabular}[t]{lllllllll}
\toprule
Question & Model & 0–45° & 45–90° & 90–135° & 135–180° & $\chi^2$ & df & p\\
\midrule
 & GPT-4o-mini & 0.819 & 0.511 & 0.209 & 0.074 & 787.19 & 3 & $<$ .001\\

 & GPT-4o & 0.315 & 0.281 & 0.340 & 0.539 & 131.16 & 3 & $<$ .001\\

 & o4-mini & 0.448 & 0.281 & 0.323 & 0.572 & 160.43 & 3 & $<$ .001\\

\multirow[t]{-4}{*}{\raggedright\arraybackslash Visual} & o3 & 0.597 & 0.368 & 0.331 & 0.721 & 304.35 & 3 & $<$ .001\\

 & GPT-4o-mini & 0.715 & 0.532 & 0.430 & 0.250 & 322.21 & 3 & $<$ .001\\

 & GPT-4o & 0.779 & 0.593 & 0.444 & 0.191 & 493.58 & 3 & $<$ .001\\

 & o4-mini & 0.865 & 0.665 & 0.629 & 0.798 & 134.73 & 3 & $<$ .001\\

\multirow[t]{-4}{*}{\raggedright\arraybackslash Spatial} & o3 & 0.609 & 0.564 & 0.499 & 0.657 & 42.38 & 3 & $<$ .001***\\
\bottomrule
\multicolumn{9}{l}{\textsuperscript{} Values are estimated marginal probabilities of correct responses.}\\
\multicolumn{9}{l}{\textsuperscript{} Simple main effects test angular disparity within each Model $\times$ Question Type combination.}\\
\end{tabular}}
\end{table}

As in Test 1, we observed significant main effects of Model, Question Type, and Angular Disparity, as well as significant interactions among these factors (all $p < .001$; Table~\ref{tab:vpt_model2_summary}), indicating that the impact of Angular Disparity varied across models and question types. For GPT-4o-mini, visual question accuracy declined sharply with increasing disparity, dropping from well above chance at low disparity (0–45°: $M = .82$) to near floor at high disparity (135–180°: $M = .07$; $\chi^2(3) = 787.19$, $p < .001$), consistent with systematic failures to adopt the figure’s visual perspective. A similar monotonic decline was observed for spatial questions, though performance differences were less severe (0–45°: $M = .72$; 135–180°: $M = .25$).

GPT-4o exhibited a qualitatively different pattern. For spatial questions, accuracy decreased steadily with Angular Disparity (0–45°: $M = .78$; 135–180°: $M = .19$; $\chi^2(3) = 493.58$, $p < .001$). In contrast, visual question performance was below chance when perspectives were shared (0–45°: $M = .32$) and increased when perspectives were inverted (135–180°: $M = .54$; $\chi^2(3) = 131.16$, $p < .001$), suggesting inconsistent or non-robust attempts to transform the symbol’s appearance.

The reasoning-oriented models o3 and o4-mini showed peak performance at low and high Angular Disparity for visual questions, with reduced accuracy at intermediate rotations. For o3, accuracy was higher when perspectives were aligned (0–45°: $M = .60$) or inverted (135–180°: $M = .72$) than at intermediate disparities (45–90°: $M = .37$; 90–135°: $M = .33$; $\chi^2(3) = 304.35$, $p < .001$). A similar U-shaped pattern was observed for o4-mini (0–45°: $M = .45$; 135–180°: $M = .57$), indicating that successful visual perspective-taking was largely confined to aligned or fully inverted viewpoints and did not generalize to intermediate rotations. For spatial questions, both models again showed U-shaped performance profiles, though accuracy was comparatively stable across disparities. This robustness was most pronounced for o4-mini, which maintained above-chance performance across all Angular Disparity bins (range $M = .63$–.87; $\chi^2(3) = 134.73$, $p < .001$).

\subsubsection{Test 3: Integrated visual and spatial PT}

In Test 3, two symbols were positioned within the figure’s field of view, slightly to the left and right (Figure~\ref{fig:vspt_example}). Unlike the previous tests, this condition involved a single visuospatial question probing how the symbol on the figure’s left or right side appeared to them, thereby jointly engaging visual and spatial perspective-taking.

Because only one visuospatial question was used, performance in Test 3 was analysed using an AI Model $\times$ Angular Disparity design. The analysis revealed a large main effect of Angular Disparity, but no significant main effect of Model, indicating broadly comparable performance across MLMs (Table~\ref{tab:vpt_model3_summary}). A significant Model $\times$ Angular Disparity interaction was also observed, reflecting differences in the magnitude of performance decline across models.

Inspection of the estimated marginal means (Table~\ref{tab:vpt_model3_emm}) and Figure~\ref{fig:test_rotation_combined} shows that all models exhibited steep declines in accuracy with increasing Angular Disparity. For example, GPT-4o-mini accuracy dropped from well above chance when perspectives were aligned (0–45°: $M = .75$) to near floor at high disparity (135–180°: $M = .02$; $\chi^2(3) = 680.52$, $p < .001$), with a similar pattern observed for GPT-4o (0–45°: $M = .75$; 135–180°: $M = .02$). The reasoning-oriented models o3 and o4-mini were less severely affected, retaining higher accuracy at intermediate and high disparities (e.g., o3: 45–90° $M = .52$, 135–180° $M = .31$; o4-mini: 45–90° $M = .47$, 135–180° $M = .28$), though performance for both models nonetheless declined to approximately chance level (25\%) at the largest disparities. Together, these results indicate a general failure to adopt visuospatial perspective in this task, even among models that showed relative robustness in earlier tests.

\begin{table}[!h]
\centering
\caption{\label{tab:vpt_model3_emm}Mean accuracy in Test 3 by Angular Disparity}
\centering
\resizebox{\ifdim\width>\linewidth\linewidth\else\width\fi}{!}{
\begin{tabular}[t]{llllllll}
\toprule
Model & 0–45° & 45–90° & 90–135° & 135–180° & $\chi^2$ & df & p\\
\midrule
GPT-4o-mini & 0.746 & 0.364 & 0.121 & 0.018 & 680.52 & 3 & $<$ .001\\
GPT-4o & 0.746 & 0.418 & 0.174 & 0.023 & 634.71 & 3 & $<$ .001\\
o4-mini & 0.796 & 0.472 & 0.253 & 0.276 & 493.27 & 3 & $<$ .001\\
o3 & 0.766 & 0.517 & 0.269 & 0.311 & 420.33 & 3 & $<$ .001\\
\bottomrule
\multicolumn{8}{l}{\textsuperscript{} Values are estimated marginal probabilities of correct responses.}\\
\multicolumn{8}{l}{\textsuperscript{} Simple main effects test angular disparity within each Model.}\\
\end{tabular}}
\end{table}

\begin{table*}[h]
\centering
\caption{\label{tab:vpt_model1_summary}Mixed effects logistic regression for Test 1: AI Model $\times$ Angular Disparity $\times$ Question Type}
\centering
\resizebox{0.9\textwidth}{!}{
\begin{tabular}[t]{lllll}
\toprule
Term & Est./($\chi^2$) & SE/(df) & z & p\\
\midrule
\textbf{A. Model Coefficients} &  &  &  & \\
Intercept & -0.403 & 0.073 & -5.487 & $<$ 0.001***\\
Model $\times$  GPT-4o & 1.246 & 0.108 & 11.582 & $<$ 0.001***\\
Model $\times$  o4-mini & 4.629 & 0.309 & 14.996 & $<$ 0.001***\\
Model $\times$  o3 & 1.617 & 0.113 & 14.285 & $<$ 0.001***\\
Question Type $\times$  Spatial & 0.683 & 0.103 & 6.597 & $<$ 0.001***\\
Angular Disparity $\times$  67.5° & 0.811 & 0.104 & 7.776 & $<$ 0.001***\\
Angular Disparity $\times$  112.5° & 1.167 & 0.107 & 10.858 & $<$ 0.001***\\
Angular Disparity $\times$  157.5° & 1.473 & 0.112 & 13.137 & $<$ 0.001***\\
Model $\times$  GPT-4o $\times$ Question Type $\times$  Spatial & -0.212 & 0.157 & -1.353 & 0.176\\
Model $\times$  o4-mini $\times$ Question Type $\times$  Spatial & -0.225 & 0.486 & -0.464 & 0.643\\
Model $\times$  o3 $\times$ Question Type $\times$  Spatial & -0.866 & 0.158 & -5.483 & $<$ 0.001***\\
Model $\times$  GPT-4o $\times$ Angular Disparity $\times$  67.5° & -0.936 & 0.152 & -6.172 & $<$ 0.001***\\
Model $\times$  o4-mini $\times$ Angular Disparity $\times$  67.5° & -3.573 & 0.330 & -10.815 & $<$ 0.001***\\
Model $\times$  o3 $\times$ Angular Disparity $\times$  67.5° & -0.686 & 0.162 & -4.225 & $<$ 0.001***\\
Model $\times$  GPT-4o $\times$ Angular Disparity $\times$  112.5° & -0.987 & 0.157 & -6.306 & $<$ 0.001***\\
Model $\times$  o4-mini $\times$ Angular Disparity $\times$  112.5° & -3.809 & 0.333 & -11.440 & $<$ 0.001***\\
Model $\times$  o3 $\times$ Angular Disparity $\times$  112.5° & -1.165 & 0.163 & -7.146 & $<$ 0.001***\\
Model $\times$  GPT-4o $\times$ Angular Disparity $\times$  157.5° & -1.056 & 0.163 & -6.463 & $<$ 0.001***\\
Model $\times$  o4-mini $\times$ Angular Disparity $\times$  157.5° & -1.903 & 0.409 & -4.659 & $<$ 0.001***\\
Model $\times$  o3 $\times$ Angular Disparity $\times$  157.5° & -1.580 & 0.165 & -9.577 & $<$ 0.001***\\
Question Type $\times$  Spatial $\times$ Angular Disparity $\times$  67.5° & -0.894 & 0.147 & -6.095 & $<$ 0.001***\\
Question Type $\times$  Spatial $\times$ Angular Disparity $\times$  112.5° & -1.377 & 0.149 & -9.244 & $<$ 0.001***\\
Question Type $\times$  Spatial $\times$ Angular Disparity $\times$  157.5° & -1.653 & 0.153 & -10.816 & $<$ 0.001***\\
Model $\times$  GPT-4o $\times$ Question Type $\times$  Spatial $\times$ Angular Disparity $\times$  67.5° & 0.649 & 0.219 & 2.965 & 0.003**\\
Model $\times$  o4-mini $\times$ Question Type $\times$  Spatial $\times$ Angular Disparity $\times$  67.5° & 0.517 & 0.514 & 1.006 & 0.314\\
Model $\times$  o3 $\times$ Question Type $\times$  Spatial $\times$ Angular Disparity $\times$  67.5° & 0.866 & 0.226 & 3.831 & $<$ 0.001***\\
Model $\times$  GPT-4o $\times$ Question Type $\times$  Spatial $\times$ Angular Disparity $\times$  112.5° & 0.717 & 0.222 & 3.234 & 0.001**\\
Model $\times$  o4-mini $\times$ Question Type $\times$  Spatial $\times$ Angular Disparity $\times$  112.5° & 0.968 & 0.516 & 1.874 & 0.061.\\
Model $\times$  o3 $\times$ Question Type $\times$  Spatial $\times$ Angular Disparity $\times$  112.5° & 1.451 & 0.226 & 6.407 & $<$ 0.001***\\
Model $\times$  GPT-4o $\times$ Question Type $\times$  Spatial $\times$ Angular Disparity $\times$  157.5° & 1.022 & 0.229 & 4.460 & $<$ 0.001***\\
Model $\times$  o4-mini $\times$ Question Type $\times$  Spatial $\times$ Angular Disparity $\times$  157.5° & 0.967 & 0.605 & 1.600 & 0.110\\
Model $\times$  o3 $\times$ Question Type $\times$  Spatial $\times$ Angular Disparity $\times$  157.5° & 1.793 & 0.228 & 7.862 & $<$ 0.001***\\
 &  &  &  & \\
\textbf{B. Type III Wald Tests} &  &  &  & \\
Intercept & 30.110 & 1 &  & $<$ 0.001***\\
Model & 393.371 & 3 &  & $<$ 0.001***\\
Question Type & 43.518 & 1 &  & $<$ 0.001***\\
Angular Disparity & 203.585 & 3 &  & $<$ 0.001***\\
Model $\times$ Question Type & 31.355 & 3 &  & $<$ 0.001***\\
Model $\times$ Angular Disparity & 274.846 & 9 &  & $<$ 0.001***\\
Question Type $\times$ Angular Disparity & 139.557 & 3 &  & $<$ 0.001***\\
Model $\times$ Question Type $\times$ Angular Disparity & 75.818 & 9 &  & $<$ 0.001***\\
\bottomrule
\multicolumn{5}{l}{\textsuperscript{} Panel A: Coefficient estimates with Wald z-tests for individual parameters}\\
\multicolumn{5}{l}{\textsuperscript{} Panel B: Type III Wald $\chi^2$ tests for omnibus effects of factors}\\
\multicolumn{5}{l}{\textsuperscript{} Significance codes: *** p $<$ 0.001, ** p $<$ 0.01, * p $<$ 0.05, . p $<$ 0.1}\\
\multicolumn{5}{l}{\textsuperscript{} Random effect variance (trial ID): 0.000}\\
\end{tabular}}
\end{table*}

\begin{table*}[h]
\centering
\caption{\label{tab:vpt_model2_summary}Mixed effects logistic regression for Test 2: AI Model $\times$ Angular Disparity $\times$ Question Type}
\centering
\resizebox{\ifdim\width>\linewidth\linewidth\else\width\fi}{!}{
\begin{tabular}[t]{lllll}
\toprule
Term & Est./($\chi^2$) & SE/(df) & z & p\\
\midrule
\textbf{A. Model Coefficients} &  &  &  & \\
Intercept & 1.510 & 0.096 & 15.787 & $<$ 0.001***\\
Model $\times$  GPT-4o & -2.287 & 0.125 & -18.318 & $<$ 0.001***\\
Model $\times$  o4-mini & -1.721 & 0.121 & -14.176 & $<$ 0.001***\\
Model $\times$  o3 & -1.118 & 0.121 & -9.219 & $<$ 0.001***\\
Question Type $\times$  Spatial & -0.588 & 0.126 & -4.678 & $<$ 0.001***\\
Angular Disparity $\times$  67.5° & -1.466 & 0.121 & -12.106 & $<$ 0.001***\\
Angular Disparity $\times$  112.5° & -2.840 & 0.131 & -21.632 & $<$ 0.001***\\
Angular Disparity $\times$  157.5° & -4.040 & 0.165 & -24.526 & $<$ 0.001***\\
Model $\times$  GPT-4o $\times$ Question Type $\times$  Spatial & 2.625 & 0.174 & 15.078 & $<$ 0.001***\\
Model $\times$  o4-mini $\times$ Question Type $\times$  Spatial & 2.653 & 0.182 & 14.558 & $<$ 0.001***\\
Model $\times$  o3 $\times$ Question Type $\times$  Spatial & 0.640 & 0.164 & 3.896 & $<$ 0.001***\\
Model $\times$  GPT-4o $\times$ Angular Disparity $\times$  67.5° & 1.304 & 0.167 & 7.800 & $<$ 0.001***\\
Model $\times$  o4-mini $\times$ Angular Disparity $\times$  67.5° & 0.738 & 0.165 & 4.484 & $<$ 0.001***\\
Model $\times$  o3 $\times$ Angular Disparity $\times$  67.5° & 0.535 & 0.162 & 3.309 & $<$ 0.001***\\
Model $\times$  GPT-4o $\times$ Angular Disparity $\times$  112.5° & 2.954 & 0.172 & 17.139 & $<$ 0.001***\\
Model $\times$  o4-mini $\times$ Angular Disparity $\times$  112.5° & 2.308 & 0.170 & 13.558 & $<$ 0.001***\\
Model $\times$  o3 $\times$ Angular Disparity $\times$  112.5° & 1.743 & 0.170 & 10.267 & $<$ 0.001***\\
Model $\times$  GPT-4o $\times$ Angular Disparity $\times$  157.5° & 4.972 & 0.196 & 25.317 & $<$ 0.001***\\
Model $\times$  o4-mini $\times$ Angular Disparity $\times$  157.5° & 4.540 & 0.194 & 23.350 & $<$ 0.001***\\
Model $\times$  o3 $\times$ Angular Disparity $\times$  157.5° & 4.596 & 0.197 & 23.375 & $<$ 0.001***\\
Question Type $\times$  Spatial $\times$ Angular Disparity $\times$  67.5° & 0.671 & 0.164 & 4.099 & $<$ 0.001***\\
Question Type $\times$  Spatial $\times$ Angular Disparity $\times$  112.5° & 1.638 & 0.171 & 9.564 & $<$ 0.001***\\
Question Type $\times$  Spatial $\times$ Angular Disparity $\times$  157.5° & 2.017 & 0.201 & 10.041 & $<$ 0.001***\\
Model $\times$  GPT-4o $\times$ Question Type $\times$  Spatial $\times$ Angular Disparity $\times$  67.5° & -1.394 & 0.232 & -6.003 & $<$ 0.001***\\
Model $\times$  o4-mini $\times$ Question Type $\times$  Spatial $\times$ Angular Disparity $\times$  67.5° & -1.113 & 0.239 & -4.651 & $<$ 0.001***\\
Model $\times$  o3 $\times$ Question Type $\times$  Spatial $\times$ Angular Disparity $\times$  67.5° & 0.072 & 0.222 & 0.324 & 0.746\\
Model $\times$  GPT-4o $\times$ Question Type $\times$  Spatial $\times$ Angular Disparity $\times$  112.5° & -3.237 & 0.235 & -13.757 & $<$ 0.001***\\
Model $\times$  o4-mini $\times$ Question Type $\times$  Spatial $\times$ Angular Disparity $\times$  112.5° & -2.434 & 0.242 & -10.043 & $<$ 0.001***\\
Model $\times$  o3 $\times$ Question Type $\times$  Spatial $\times$ Angular Disparity $\times$  112.5° & -0.987 & 0.228 & -4.336 & $<$ 0.001***\\
Model $\times$  GPT-4o $\times$ Question Type $\times$  Spatial $\times$ Angular Disparity $\times$  157.5° & -5.656 & 0.261 & -21.704 & $<$ 0.001***\\
Model $\times$  o4-mini $\times$ Question Type $\times$  Spatial $\times$ Angular Disparity $\times$  157.5° & -2.997 & 0.266 & -11.274 & $<$ 0.001***\\
Model $\times$  o3 $\times$ Question Type $\times$  Spatial $\times$ Angular Disparity $\times$  157.5° & -2.368 & 0.251 & -9.446 & $<$ 0.001***\\
 &  &  &  & \\
\textbf{B. Type III Wald Tests} &  &  &  & \\
Intercept & 249.222 & 1 &  & $<$ 0.001***\\
Model & 367.700 & 3 &  & $<$ 0.001***\\
Question Type & 21.884 & 1 &  & $<$ 0.001***\\
Angular Disparity & 787.189 & 3 &  & $<$ 0.001***\\
Model $\times$ Question Type & 366.448 & 3 &  & $<$ 0.001***\\
Model $\times$ Angular Disparity & 906.376 & 9 &  & $<$ 0.001***\\
Question Type $\times$ Angular Disparity & 145.116 & 3 &  & $<$ 0.001***\\
Model $\times$ Question Type $\times$ Angular Disparity & 614.692 & 9 &  & $<$ 0.001***\\
\bottomrule
\multicolumn{5}{l}{\textsuperscript{} Panel A: Coefficient estimates with Wald z-tests for individual parameters}\\
\multicolumn{5}{l}{\textsuperscript{} Panel B: Type III Wald $\chi^2$ tests for omnibus effects of factors}\\
\multicolumn{5}{l}{\textsuperscript{} Significance codes: *** p $<$ 0.001, ** p $<$ 0.01, * p $<$ 0.05, . p $<$ 0.1}\\
\multicolumn{5}{l}{\textsuperscript{} Random effect variance (trial ID): 0.000}\\
\end{tabular}}
\end{table*}
\hspace{10pt}

\begin{table*}[h]
\centering
\caption{\label{tab:vpt_model3_summary}Mixed effects logistic regression for Test 3: AI Model $\times$ Angular Disparity}
\centering
\resizebox{\ifdim\width>\linewidth\linewidth\else\width\fi}{!}{
\begin{tabular}[t]{lllll}
\toprule
Term & Est./($\chi^2$) & SE/(df) & z & p\\
\midrule
\textbf{A. Model Coefficients} &  &  &  & \\
Intercept & 1.076 & 0.084 & 12.860 & $<$ 0.001***\\
Model $\times$  GPT-4o & 0.000 & 0.118 & 0.000 & 1.000\\
Model $\times$  o4-mini & 0.287 & 0.123 & 2.332 & 0.020*\\
Model $\times$  o3 & 0.108 & 0.120 & 0.902 & 0.367\\
Angular Disparity $\times$  67.5° & -1.636 & 0.112 & -14.588 & $<$ 0.001***\\
Angular Disparity $\times$  112.5° & -3.059 & 0.141 & -21.775 & $<$ 0.001***\\
Angular Disparity $\times$  157.5° & -5.100 & 0.292 & -17.468 & $<$ 0.001***\\
Model $\times$  GPT-4o $\times$ Angular Disparity $\times$  67.5° & 0.229 & 0.158 & 1.448 & 0.148\\
Model $\times$  o4-mini $\times$ Angular Disparity $\times$  67.5° & 0.161 & 0.161 & 1.001 & 0.317\\
Model $\times$  o3 $\times$ Angular Disparity $\times$  67.5° & 0.522 & 0.159 & 3.291 & $<$ 0.001***\\
Model $\times$  GPT-4o $\times$ Angular Disparity $\times$  112.5° & 0.426 & 0.190 & 2.236 & 0.025*\\
Model $\times$  o4-mini $\times$ Angular Disparity $\times$  112.5° & 0.612 & 0.187 & 3.265 & 0.001**\\
Model $\times$  o3 $\times$ Angular Disparity $\times$  112.5° & 0.876 & 0.184 & 4.750 & $<$ 0.001***\\
Model $\times$  GPT-4o $\times$ Angular Disparity $\times$  157.5° & 0.274 & 0.391 & 0.701 & 0.483\\
Model $\times$  o4-mini $\times$ Angular Disparity $\times$  157.5° & 2.770 & 0.317 & 8.752 & $<$ 0.001***\\
Model $\times$  o3 $\times$ Angular Disparity $\times$  157.5° & 3.119 & 0.314 & 9.920 & $<$ 0.001***\\
 &  &  &  & \\
\textbf{B. Type III Wald Tests} &  &  &  & \\
Intercept & 165.368 & 1 &  & $<$ 0.001***\\
Model & 7.110 & 3 &  & 0.068.\\
Angular Disparity & 680.520 & 3 &  & $<$ 0.001***\\
Model $\times$ Angular Disparity & 195.476 & 9 &  & $<$ 0.001***\\
\bottomrule
\multicolumn{5}{l}{\textsuperscript{} Panel A: Coefficient estimates with Wald z-tests for individual parameters}\\
\multicolumn{5}{l}{\textsuperscript{} Panel B: Type III Wald $\chi^2$ tests for omnibus effects of factors}\\
\multicolumn{5}{l}{\textsuperscript{} Significance codes: *** p $<$ 0.001, ** p $<$ 0.01, * p $<$ 0.05, . p $<$ 0.1}\\
\multicolumn{5}{l}{\textsuperscript{} Random effect variance (trial ID): 0.000}\\
\end{tabular}}
\end{table*}

\subsection{Director Task}
\label{appendix:supp_director_analysis}
We analysed MLM performance on the \textit{Director Task} using mixed-effects logistic regression models. Main and interaction effects were evaluated with Type III Wald $\chi^2$ tests, and condition differences were examined via post-hoc comparisons of estimated marginal means.

\begin{figure}[!h]
    \hspace{1.40cm}
    \begin{center} 
    \centering
    \includegraphics[width=0.6\linewidth]{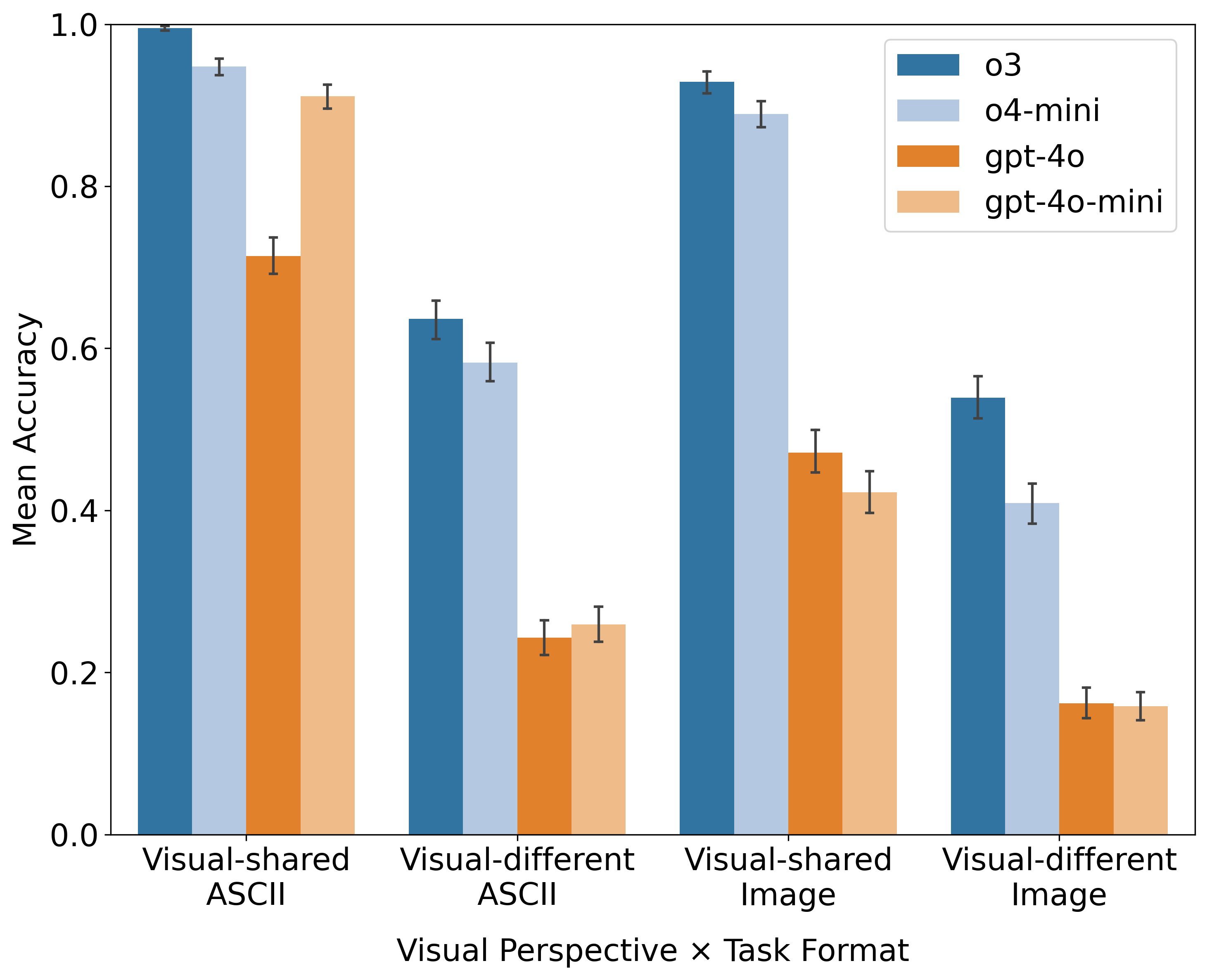}
    \caption{MLM mean accuracy in the \textit{Director Task} for Visual Perspective conditions (Visual-shared and Visual-different) across Task Formats (ASCII and Image). Spatial-different and participant POV trials were excluded. Error bars are 95 percent confidence intervals.}
    \label{fig:visualxformat}
    \end{center}
\end{figure}

We first assessed Level~2 \textit{visual} perspective taking (visual PT) as a function of task format (image vs. ASCII). To isolate visual PT, we excluded spatial-different trials and restricted the analysis to trials presented from the director’s point of view. Mean accuracies are shown in Figure \ref{fig:visualxformat}, with model summaries in Tables \ref{tab:visualxformat_model} and \ref{tab:visualxformat_pairwise}. Overall performance differed substantially across models ($\chi^2$(1) = 498.04, $p$ < .001). Accuracy was lower in the image-based task ($M$  = 0.52) than in the ASCII version ($M$ = 0.74, $\chi^2$(1) = 88.35, $p$ < .001), indicating that visual processing demands impaired performance. Crucially, MLMs showed a pronounced deficit in visual perspective taking: accuracy dropped sharply from visual-shared trials ($M$ = 0.88) to visual-different trials ($M$ = 0.35; $\chi^2$(3) = 141.27, $p$ < .001). While reasoning models' (o3, o4-mini) baseline performance was more robust across task formats, the effect of visual perspective was large for all models (all $log(OR)$ > 6, all $p$ < .001). Notably, o3 performed near ceiling on visual-shared trials ($M$ = 0.98) yet showed a collapse in accuracy on visual-different trials ($M$ = 0.59; $log(OR)$ = 38.67, $p$ < .001), indicating a persistent failure to inhibit privileged visual information and adopt the director's perspective.

We next examined whether the use of different relative adjectives modulated visual perspective taking performance (Tables \ref{tab:visualxadjective_model} and \ref{tab:visualxadjective_pairwise}). To assess general effects of relative adjectives independent of spatial perspective taking, we again excluded spatial-different trials and restricted the analysis to trials presented from the director’s POV, and fit an AI Model $\times$ Relative Adjective (None, Size, Spatial) $\times$ Visual Perspective (Visual-shared, Visual-different) model, with Task Format as a covariate. This analysis revealed no main effect of Relative Adjective ($\chi^2$(2) = 2.69, $p$ = .260), indicating that baseline performance did not differ across relative adjective conditions. We did find a weak interaction with Visual Perspective ($\chi^2$(2) = 6.25, $p$ = .044), though the effect of Visual Perspective was large in all relative adjective conditions. A large AI Model $\times$ Relative Adjective interaction ($\chi^2$(1) = 182.46, $p$ < .001) further underscored baseline capability differences across MLMs, with reasoning models (o3, o4-mini) showing consistently high accuracy across relative adjective conditions when visual-perspective-taking was not required (all $M$ > 0.89). 

\begin{figure}[!h]
    \hspace{1.40cm}
    \begin{center} 
    \centering
    \includegraphics[width=0.6\linewidth]{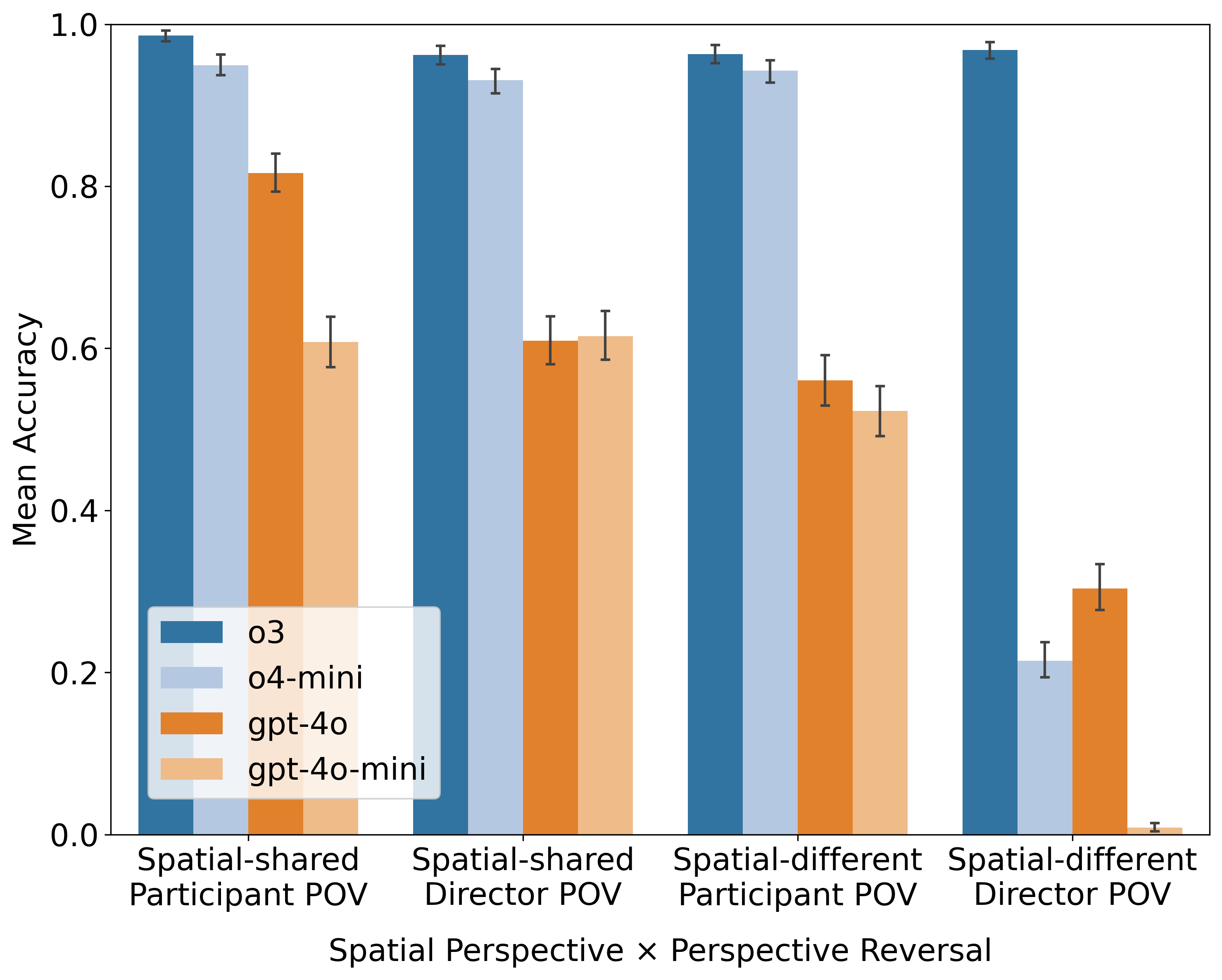}
    \caption{MLM mean accuracy in the \textit{Director Task} on trials with spatial adjectives. Spatial perspective taking condition levels are Spatial-shared (Vertical:'topmost'/'bottommost') and Spatial-different (Horizontal: 'leftmost'/'rightmost'). Perspective Reversal condition levels are Participant point-of-view (POV; 'from your point of view') and Director POV ('from my point of view'). The participant is required to adopt the director's spatial perspective on spatial-different trials from the director's POV only (rightmost grouping). Visual-different perspective-taking trials are not included. Error bars are 95 percent confidence intervals.}
    \label{fig:spatialxreversal}
    \end{center}
\end{figure}

Following that result, we isolated Level~2 \textit{spatial} perspective taking (spatial PT), by restricting our analyses to visual-shared trials containing spatial adjectives only and fit an AI Model $\times$ Spatial Perspective (Shared (i.e., vertical adjective) vs. Different (i.e., horizontal adjective)) $\times$ Perspective Reversal (Participant POV vs. Director POV) model, again controlling for Task Format (Tables \ref{tab:spatialxreversal_model}, \ref{tab:spatialxreversal_pairwise}). As target items specified using vertical relative adjectives (e.g., topmost, bottommost) are the same regardless of whether the director specifies ``from my point-of-view'' or ``from your point-of-view'', the participant is required to adopt the director's spatial perspective on spatial-different (i.e., horizontal adjective) trials from the director's point-of-view only. We thus found no main effect of Perspective Reversal ($\chi^2$(1) = 0.02, $p$ = .888), but a strong Spatial Perspective $\times$ Perspective Reversal interaction ($\chi^2$(1) = 120.93, $p$ < .001), which reflected robust performance across all conditions (all $M$ > 0.84) aside from spatial-different trials from the director's POV ($M$ = 0.28). We also found a three-way interaction ($\chi^2$(3) = 361.378, $p$ < .001), suggesting differences across MLMs. Post-hoc comparisons indicated substantial spatial PT impairments for GPT-4o-mini ($log(OR)$ = 224.27, $p$ < .001) and o4-mini ($log(OR)$ = 65.30, $p$ < .001). GPT-4o showed poorer performance on horizontal trials regardless of perspective reversal (both $p$ < .001), while o3's accuracy was not affected by the spatial perspective taking manipulation (all $M$ > 0.96, $log(OR)$ = 0.89, $p$ = 0.698; see Figure~\ref{fig:spatialxreversal}).

We then investigated the interaction between visual and spatial Level~2 perspective taking demands by fitting an AI Model $\times$ Visual PT $\times$ Spatial PT mixed-effects model. The findings from this analysis are summarised in Tables~\ref{tab:visualxspatial_model},~\ref{tab:visualxspatial_pairwise}, and are described in the main text. Figure~\ref{fig:visualxspatial_divided}, displays the spatial PT and visual PT accuracy scores separately for image and ASCII tasks.

\begin{table*}[!h]
\centering
\caption{\label{tab:visualxformat_pairwise}Pairwise comparisons with estimated marginal means: AI Model $\times$ Visual PT $\times$ Task Format}
\centering
\resizebox{1\textwidth}{!}{
\begin{tabular}[t]{llllllll}
\toprule
Contrast & Context & \multicolumn{2}{l}{Estimated marginal means} & Estimate & SE & z & p \\
\midrule
\textbf{A. Perspective $\times$ Format} &  &  &  &  &  &  & \\
Shared vs. Different & ASCII & Shared: 0.951 & Different: 0.420 & 26.881 & 7.766 & 11.393 & $<$ 0.001***\\
Shared vs. Different & Image & Shared: 0.743 & Different: 0.292 & 6.999 & 1.934 & 7.041 & $<$ 0.001***\\
 &  &  &  &  &  &  & \\
\textbf{B. Format $\times$ Perspective} &  &  &  &  &  &  & \\
ASCII vs. Image & Visual-shared & ASCII: 0.951 & Image: 0.743 & 6.730 & 1.952 & 6.574 & $<$ 0.001***\\
ASCII vs. Image & Visual-different & ASCII: 0.420 & Image: 0.292 & 1.752 & 0.482 & 2.038 & 0.042*\\
 &  &  &  &  &  &  & \\
\textbf{C. Perspective $\times$ Model} &  &  &  &  &  &  & \\
Shared vs. Different & gpt-4o-mini & Shared: 0.741 & Different: 0.202 & 11.295 & 2.315 & 11.828 & $<$ 0.001***\\
Shared vs. Different & gpt-4o & Shared: 0.605 & Different: 0.197 & 6.225 & 1.256 & 9.060 & $<$ 0.001***\\
Shared vs. Different & o4-mini & Shared: 0.928 & Different: 0.497 & 13.021 & 2.693 & 12.408 & $<$ 0.001***\\
Shared vs. Different & o3 & Shared: 0.982 & Different: 0.591 & 38.665 & 10.143 & 13.932 & $<$ 0.001***\\
 &  &  &  &  &  &  & \\
\textbf{D. Format $\times$ Model} &  &  &  &  &  &  & \\
ASCII vs. Image & gpt-4o-mini & ASCII: 0.664 & Image: 0.269 & 5.375 & 1.102 & 8.204 & $<$ 0.001***\\
ASCII vs. Image & gpt-4o & ASCII: 0.478 & Image: 0.292 & 2.220 & 0.448 & 3.951 & $<$ 0.001***\\
ASCII vs. Image & o4-mini & ASCII: 0.843 & Image: 0.703 & 2.257 & 0.467 & 3.936 & $<$ 0.001***\\
ASCII vs. Image & o3 & ASCII: 0.953 & Image: 0.798 & 5.162 & 1.354 & 6.257 & $<$ 0.001***\\
\bottomrule
\multicolumn{8}{l}{\textsuperscript{} Estimate is the the log-odds ratio between conditions.}\\
\multicolumn{8}{l}{\textsuperscript{} P-values adjusted using the Bonferroni method within each family of comparisons.}\\
\multicolumn{8}{l}{\textsuperscript{} Significance codes: *** p $<$ 0.001, ** p $<$ 0.01, * p $<$ 0.05, . p $<$ 0.1}\\
\end{tabular}}
\end{table*}

\begin{table*}[ht]
\centering
\caption{\label{tab:visualxformat_model}Mixed effects logistic regression: AI Model $\times$ Visual PT $\times$ Task Format}
\centering
\resizebox{0.9\textwidth}{!}{
\begin{tabular}[t]{lllll}
\toprule
Term & Est./($\chi^2$) & SE/(df) & z & p\\
\midrule
\textbf{A. Model Coefficients} &  &  &  & \\
Intercept & 2.427 & 0.214 & 11.352 & $<$ 0.001***\\
Model: GPT-4o & -1.455 & 0.109 & -13.394 & $<$ 0.001***\\
Model: o4-mini & 0.585 & 0.139 & 4.222 & $<$ 0.001***\\
Model: o3 & 3.031 & 0.346 & 8.752 & $<$ 0.001***\\
Format: Image & -2.748 & 0.292 & -9.400 & $<$ 0.001***\\
Visual PT: Different & -3.490 & 0.294 & -11.886 & $<$ 0.001***\\
Model: GPT-4o $\times$ Format: Image & 1.657 & 0.131 & 12.610 & $<$ 0.001***\\
Model: o4-mini $\times$ Format: Image & 1.831 & 0.170 & 10.801 & $<$ 0.001***\\
Model: o3 $\times$ Format: Image & -0.124 & 0.364 & -0.339 & 0.735\\
Model: GPT-4o $\times$ Visual PT: Different & 1.368 & 0.138 & 9.918 & $<$ 0.001***\\
Model: o4-mini $\times$ Visual PT: Different & 0.822 & 0.158 & 5.196 & $<$ 0.001***\\
Model: o3 $\times$ Visual PT: Different & -1.394 & 0.355 & -3.930 & $<$ 0.001***\\
Format: Image $\times$ Visual PT: Different & 2.132 & 0.410 & 5.200 & $<$ 0.001***\\
Model: GPT-4o $\times$ Format: Image $\times$ Visual PT: Different & -1.545 & 0.186 & -8.314 & $<$ 0.001***\\
Model: o4-mini $\times$ Format: Image $\times$ Visual PT: Different & -1.928 & 0.206 & -9.358 & $<$ 0.001***\\
Model: o3 $\times$ Format: Image $\times$ Visual PT: Different & 0.328 & 0.383 & 0.856 & 0.392\\
 &  &  &  & \\
\textbf{B. Type III Wald Tests} &  &  &  & \\
Intercept & 128.877 & 1 &  & $<$ 0.001***\\
Model & 498.038 & 3 &  & $<$ 0.001***\\
Format & 88.354 & 1 &  & $<$ 0.001***\\
Visual PT & 141.269 & 1 &  & $<$ 0.001***\\
Model $\times$ Format & 200.406 & 3 &  & $<$ 0.001***\\
Model $\times$ Visual PT & 141.104 & 3 &  & $<$ 0.001***\\
Format $\times$ Visual PT & 27.044 & 1 &  & $<$ 0.001***\\
Model $\times$ Format $\times$ Visual PT & 121.048 & 3 &  & $<$ 0.001***\\
\bottomrule
\multicolumn{5}{l}{\textsuperscript{} Panel A: Coefficient estimates with Wald z-tests for individual parameters}\\
\multicolumn{5}{l}{\textsuperscript{} Panel B: Type III Wald $\chi^2$ tests for omnibus effects of factors}\\
\multicolumn{5}{l}{\textsuperscript{} Significance codes: *** p $<$ 0.001, ** p $<$ 0.01, * p $<$ 0.05, . p $<$ 0.1}\\
\multicolumn{5}{l}{\textsuperscript{} Random effect variance (Trial ID): 0.871}\\
\end{tabular}}
\end{table*}

\begin{table*}[!h]
\centering
\caption{\label{tab:visualxadjective_pairwise}Pairwise comparisons with estimated marginal means: AI Model $\times$ Visual PT $\times$ Relative Adjective}
\centering
\resizebox{0.9\textwidth}{!}{
\begin{tabular}[t]{llllllll}
\toprule
Contrast & Context & \multicolumn{2}{l}{Estimated marginal means} & Estimate & SE & z & p \\
\midrule
\textbf{A. GPT-4o-mini} &  &  &  &  &  &  & \\
Shared vs. Different & None & Shared: 0.739 & Different: 0.135 & 18.178 & 6.020 & 8.757 & $<$ 0.001***\\
Shared vs. Different & Size & Shared: 0.715 & Different: 0.282 & 6.403 & 2.090 & 5.688 & $<$ 0.001***\\
Shared vs. Different & Spatial & Shared: 0.630 & Different: 0.198 & 6.889 & 2.248 & 5.915 & $<$ 0.001***\\
 &  &  &  &  &  &  & \\
\textbf{B. GPT-4o} &  &  &  &  &  &  & \\
Shared vs. Different & None & Shared: 0.416 & Different: 0.166 & 3.594 & 1.178 & 3.903 & $<$ 0.001***\\
Shared vs. Different & Size & Shared: 0.787 & Different: 0.280 & 9.504 & 3.115 & 6.870 & $<$ 0.001***\\
Shared vs. Different & Spatial & Shared: 0.623 & Different: 0.151 & 9.318 & 3.060 & 6.795 & $<$ 0.001***\\
 &  &  &  &  &  &  & \\
\textbf{C. o4-mini} &  &  &  &  &  &  & \\
Shared vs. Different & None & Shared: 0.895 & Different: 0.600 & 5.713 & 1.885 & 5.281 & $<$ 0.001***\\
Shared vs. Different & Size & Shared: 0.965 & Different: 0.509 & 26.765 & 9.269 & 9.492 & $<$ 0.001***\\
Shared vs. Different & Spatial & Shared: 0.939 & Different: 0.389 & 24.228 & 8.172 & 9.451 & $<$ 0.001***\\
 &  &  &  &  &  &  & \\
\textbf{D. o3} &  &  &  &  &  &  & \\
Shared vs. Different & None & Shared: 0.957 & Different: 0.722 & 8.501 & 2.943 & 6.182 & $<$ 0.001***\\
Shared vs. Different & Size & Shared: 0.990 & Different: 0.535 & 86.428 & 34.709 & 11.104 & $<$ 0.001***\\
Shared vs. Different & Spatial & Shared: 0.967 & Different: 0.510 & 28.174 & 9.923 & 9.479 & $<$ 0.001***\\
\bottomrule
\multicolumn{8}{l}{\textsuperscript{} Estimate is the the log-odds ratio between conditions.}\\
\multicolumn{8}{l}{\textsuperscript{} P-values adjusted using the Bonferroni method within each family of comparisons.}\\
\multicolumn{8}{l}{\textsuperscript{} Significance codes: *** p $<$ 0.001, ** p $<$ 0.01, * p $<$ 0.05, . p $<$ 0.1}\\
\end{tabular}}
\end{table*}

\begin{table*}[!h]
\centering
\caption{\label{tab:visualxadjective_model}Mixed effects logistic regression: AI Model $\times$ Visual PT $\times$ Relative Adjective}
\centering
\resizebox{1\textwidth}{!}{
\begin{tabular}[t]{lllll}
\toprule
Term & Est./($\chi^2$) & SE/(df) & z & p\\
\midrule
\textbf{A. Model Coefficients} &  &  &  & \\
Intercept & 1.618 & 0.249 & 6.501 & $<$ 0.001***\\
Model: GPT-4o & -1.380 & 0.103 & -13.341 & $<$ 0.001***\\
Model: o4-mini & 1.106 & 0.121 & 9.103 & $<$ 0.001***\\
Model: o3 & 2.055 & 0.158 & 13.034 & $<$ 0.001***\\
Rel. Adj: Size & -0.119 & 0.327 & -0.365 & 0.715\\
Rel. Adj: Spatial & -0.510 & 0.326 & -1.567 & 0.117\\
Visual PT: Different & -2.900 & 0.331 & -8.757 & $<$ 0.001***\\
Format: Image & -1.153 & 0.182 & -6.342 & $<$ 0.001***\\
Model: GPT-4o $\times$ Rel. Adj: Size & 1.765 & 0.149 & 11.821 & $<$ 0.001***\\
Model: o4-mini $\times$ Rel. Adj: Size & 1.297 & 0.199 & 6.507 & $<$ 0.001***\\
Model: o3 $\times$ Rel. Adj: Size & 1.621 & 0.302 & 5.372 & $<$ 0.001***\\
Model: GPT-4o $\times$ Rel. Adj: Spatial & 1.351 & 0.142 & 9.508 & $<$ 0.001***\\
Model: o4-mini $\times$ Rel. Adj: Spatial & 1.097 & 0.184 & 5.964 & $<$ 0.001***\\
Model: o3 $\times$ Rel. Adj: Spatial & 0.791 & 0.233 & 3.391 & $<$ 0.001***\\
Model: GPT-4o $\times$ Visual PT: Different & 1.621 & 0.162 & 9.989 & $<$ 0.001***\\
Model: o4-mini $\times$ Visual PT: Different & 1.157 & 0.165 & 7.015 & $<$ 0.001***\\
Model: o3 $\times$ Visual PT: Different & 0.760 & 0.195 & 3.892 & $<$ 0.001***\\
Rel. Adj: Size $\times$ Visual PT: Different & 1.043 & 0.465 & 2.244 & 0.025*\\
Rel. Adj: Spatial $\times$ Visual PT: Different & 0.970 & 0.465 & 2.087 & 0.037*\\
Model: GPT-4o $\times$ Rel. Adj: Size $\times$ Visual PT: Different & -2.016 & 0.219 & -9.189 & $<$ 0.001***\\
Model: o4-mini $\times$ Rel. Adj: Size $\times$ Visual PT: Different & -2.588 & 0.247 & -10.475 & $<$ 0.001***\\
Model: o3 $\times$ Rel. Adj: Size $\times$ Visual PT: Different & -3.363 & 0.336 & -9.994 & $<$ 0.001***\\
Model: GPT-4o $\times$ Rel. Adj: Spatial $\times$ Visual PT: Different & -1.923 & 0.223 & -8.622 & $<$ 0.001***\\
Model: o4-mini $\times$ Rel. Adj: Spatial $\times$ Visual PT: Different & -2.415 & 0.237 & -10.179 & $<$ 0.001***\\
Model: o3 $\times$ Rel. Adj: Spatial $\times$ Visual PT: Different & -2.169 & 0.278 & -7.790 & $<$ 0.001***\\
 &  &  &  & \\
\textbf{B. Type III Wald Tests} &  &  &  & \\
Intercept & 42.268 & 1 &  & $<$ 0.001***\\
Model & 701.100 & 3 &  & $<$ 0.001***\\
Rel. Adj & 2.693 & 2 &  & 0.260\\
Visual PT & 76.687 & 1 &  & $<$ 0.001***\\
Format & 40.226 & 1 &  & $<$ 0.001***\\
Model $\times$ Rel. Adj & 170.216 & 6 &  & $<$ 0.001***\\
Model $\times$ Visual PT & 105.717 & 3 &  & $<$ 0.001***\\
Rel. Adj $\times$ Visual PT & 6.247 & 2 &  & 0.044*\\
Model $\times$ Rel. Adj $\times$ Visual PT & 207.900 & 6 &  & $<$ 0.001***\\
\bottomrule
\multicolumn{5}{l}{\textsuperscript{} Panel A: Coefficient estimates with Wald z-tests for individual parameters}\\
\multicolumn{5}{l}{\textsuperscript{} Panel B: Type III Wald $\chi^2$ tests for omnibus effects of factors}\\
\multicolumn{5}{l}{\textsuperscript{} Significance codes: *** p $<$ 0.001, ** p $<$ 0.01, * p $<$ 0.05, . p $<$ 0.1}\\
\multicolumn{5}{l}{\textsuperscript{} Random effect variance (Trial ID): 0.394}\\
\end{tabular}}
\end{table*}

\begin{table*}[!h]
\centering
\caption{\label{tab:spatialxreversal_pairwise}Pairwise comparisons with estimated marginal means: AI Model $\times$ Spatial PT $\times$ Perspective Reversal}
\centering
\resizebox{0.9\textwidth}{!}{
\begin{tabular}[t]{llllllll}
\toprule
Contrast & Context & \multicolumn{2}{l}{Estimated marginal means} & Estimate & SE & z & p \\
\midrule
\textbf{A. GPT-4o-mini} &  &  &  &  &  &  & \\
Shared vs. Different & Participant POV & Shared: 0.637 & Different: 0.526 & 1.586 & 0.346 & 2.112 & 0.035*\\
Shared vs. Different & Director POV & Shared: 0.630 & Different: 0.008 & 224.268 & 88.320 & 13.745 & $<$ 0.001***\\
 &  &  &  &  &  &  & \\
\textbf{B. GPT-4o} &  &  &  &  &  &  & \\
Shared vs. Different & Participant POV & Shared: 0.861 & Different: 0.568 & 4.724 & 1.067 & 6.873 & $<$ 0.001***\\
Shared vs. Different & Director POV & Shared: 0.623 & Different: 0.292 & 4.023 & 0.873 & 6.411 & $<$ 0.001***\\
 &  &  &  &  &  &  & \\
\textbf{C. o4-mini} &  &  &  &  &  &  & \\
Shared vs. Different & Participant POV & Shared: 0.963 & Different: 0.949 & 1.411 & 0.390 & 1.248 & 0.212\\
Shared vs. Different & Director POV & Shared: 0.939 & Different: 0.191 & 65.301 & 15.693 & 17.389 & $<$ 0.001***\\
 &  &  &  &  &  &  & \\
\textbf{D. o3} &  &  &  &  &  &  & \\
Shared vs. Different & Participant POV & Shared: 0.990 & Different: 0.967 & 3.361 & 1.212 & 3.361 & $<$ 0.001***\\
Shared vs. Different & Director POV & Shared: 0.967 & Different: 0.971 & 0.890 & 0.268 & -0.388 & 0.698\\
\bottomrule
\multicolumn{8}{l}{\textsuperscript{} Estimate is the the log-odds ratio between conditions.}\\
\multicolumn{8}{l}{\textsuperscript{} P-values adjusted using the Bonferroni method within each family of comparisons.}\\
\multicolumn{8}{l}{\textsuperscript{} Significance codes: *** p $<$ 0.001, ** p $<$ 0.01, * p $<$ 0.05, . p $<$ 0.1}\\
\end{tabular}}
\end{table*}

\begin{table*}[!h]
\centering
\caption{\label{tab:spatialxreversal_model}Mixed effects logistic regression: AI Model $\times$ Spatial PT $\times$ Perspective Reversal}
\centering
\resizebox{1\textwidth}{!}{
\begin{tabular}[t]{lllll}
\toprule
Term & Est./($\chi^2$) & SE/(df) & z & p\\
\midrule
\textbf{A. Model Coefficients} &  &  &  & \\
Intercept & 1.310 & 0.173 & 7.566 & $<$ 0.001***\\
Model: GPT-4o & 1.261 & 0.115 & 10.942 & $<$ 0.001***\\
Model: o4-mini & 2.705 & 0.159 & 17.053 & $<$ 0.001***\\
Model: o3 & 4.041 & 0.265 & 15.260 & $<$ 0.001***\\
Spatial PT: Different & -0.461 & 0.218 & -2.112 & 0.035*\\
Persp. Reversal: Director POV & -0.031 & 0.219 & -0.141 & 0.888\\
Format: Image & -1.493 & 0.146 & -10.256 & $<$ 0.001***\\
Model: GPT-4o $\times$ Spatial PT: Different & -1.092 & 0.151 & -7.248 & $<$ 0.001***\\
Model: o4-mini $\times$ Spatial PT: Different & 0.116 & 0.218 & 0.533 & 0.594\\
Model: o3 $\times$ Spatial PT: Different & -0.751 & 0.319 & -2.358 & 0.018*\\
Model: GPT-4o $\times$ Persp. Reversal: Director POV & -1.290 & 0.151 & -8.540 & $<$ 0.001***\\
Model: o4-mini $\times$ Persp. Reversal: Director POV & -0.500 & 0.210 & -2.375 & 0.018*\\
Model: o3 $\times$ Persp. Reversal: Director POV & -1.193 & 0.316 & -3.778 & $<$ 0.001***\\
Spatial PT: Different $\times$ Persp. Reversal: Director POV & -4.952 & 0.450 & -10.997 & $<$ 0.001***\\
Model: GPT-4o $\times$ Spatial PT: Different $\times$ Persp. Reversal: Director POV & 5.113 & 0.387 & 13.226 & $<$ 0.001***\\
Model: o4-mini $\times$ Spatial PT: Different $\times$ Persp. Reversal: Director POV & 1.118 & 0.429 & 2.603 & 0.009**\\
Model: o3 $\times$ Spatial PT: Different $\times$ Persp. Reversal: Director POV & 6.281 & 0.523 & 12.021 & $<$ 0.001***\\
 &  &  &  & \\
\textbf{B. Type III Wald Tests} &  &  &  & \\
Intercept & 57.241 & 1 &  & $<$ 0.001***\\
Model & 472.449 & 3 &  & $<$ 0.001***\\
Spatial PT & 4.461 & 1 &  & 0.035*\\
Persp. Reversal & 0.020 & 1 &  & 0.888\\
Format & 105.178 & 1 &  & $<$ 0.001***\\
Model $\times$ Spatial PT & 62.842 & 3 &  & $<$ 0.001***\\
Model $\times$ Persp. Reversal & 77.080 & 3 &  & $<$ 0.001***\\
Spatial PT $\times$ Persp. Reversal & 120.928 & 1 &  & $<$ 0.001***\\
Model $\times$ Spatial PT $\times$ Persp. Reversal & 361.378 & 3 &  & $<$ 0.001***\\
\bottomrule
\multicolumn{5}{l}{\textsuperscript{} Panel A: Coefficient estimates with Wald z-tests for individual parameters}\\
\multicolumn{5}{l}{\textsuperscript{} Panel B: Type III Wald $\chi^2$ tests for omnibus effects of factors}\\
\multicolumn{5}{l}{\textsuperscript{} Significance codes: *** p $<$ 0.001, ** p $<$ 0.01, * p $<$ 0.05, . p $<$ 0.1}\\
\multicolumn{5}{l}{\textsuperscript{} Random effect variance (Trial ID): 2.073}\\
\end{tabular}}
\end{table*}

\begin{table*}[!h]
\centering
\caption{\label{tab:visualxspatial_pairwise}Pairwise comparisons with estimated marginal means: AI Model $\times$ Visual PT $\times$ Spatial PT}
\centering
\resizebox{1\textwidth}{!}{
\begin{tabular}[t]{llllllll}
\toprule
Contrast & Context & \multicolumn{2}{l}{Estimated marginal means} & Estimate & SE & z & p \\
\midrule
\textbf{A. Visual $\times$ Spatial} &  &  &  &  &  &  & \\
Visual-shared vs. Visual-different & Spatial-shared & Shared: 0.856 & Different: 0.292 & 14.390 & 3.710 & 10.344 & $<$ 0.001***\\
Visual-shared vs. Visual-different & Spatial-different & Shared: 0.285 & Different: 0.087 & 4.183 & 1.148 & 5.212 & $<$ 0.001***\\
 &  &  &  &  &  &  & \\
\textbf{B. Spatial $\times$ Visual} &  &  &  &  &  &  & \\
Spatial-shared vs. Spatial-different & Visual-shared & Shared: 0.856 & Different: 0.285 & 14.914 & 4.073 & 9.894 & $<$ 0.001***\\
Spatial-shared vs. Spatial-different & Visual-different & Shared: 0.292 & Different: 0.087 & 4.335 & 1.124 & 5.656 & $<$ 0.001***\\
 &  &  &  &  &  &  & \\
\textbf{C. Visual $\times$ Model} &  &  &  &  &  &  & \\
Visual-shared vs. Visual-different & gpt-4o-mini & Shared: 0.103 & Different: 0.077 & 1.365 & 0.363 & 1.171 & 0.242\\
Visual-shared vs. Visual-different & gpt-4o & Shared: 0.453 & Different: 0.135 & 5.303 & 1.027 & 8.610 & $<$ 0.001***\\
Visual-shared vs. Visual-different & o4-mini & Shared: 0.657 & Different: 0.168 & 9.465 & 1.908 & 11.148 & $<$ 0.001***\\
Visual-shared vs. Visual-different & o3 & Shared: 0.969 & Different: 0.368 & 52.875 & 11.386 & 18.426 & $<$ 0.001***\\
 &  &  &  &  &  &  & \\
\textbf{D. Spatial $\times$ Model} &  &  &  &  &  &  & \\
Spatial-shared vs. Spatial-different & gpt-4o-mini & Shared: 0.393 & Different: 0.015 & 43.618 & 11.602 & 14.194 & $<$ 0.001***\\
Spatial-shared vs. Spatial-different & gpt-4o & Shared: 0.351 & Different: 0.193 & 2.270 & 0.440 & 4.230 & $<$ 0.001***\\
Spatial-shared vs. Spatial-different & o4-mini & Shared: 0.757 & Different: 0.111 & 25.043 & 5.050 & 15.972 & $<$ 0.001***\\
Spatial-shared vs. Spatial-different & o3 & Shared: 0.846 & Different: 0.765 & 1.686 & 0.363 & 2.426 & 0.015*\\
\bottomrule
\multicolumn{8}{l}{\textsuperscript{} Estimate is the the log-odds ratio between conditions.}\\
\multicolumn{8}{l}{\textsuperscript{} P-values adjusted using the Bonferroni method within each family of comparisons.}\\
\multicolumn{8}{l}{\textsuperscript{} Significance codes: *** p $<$ 0.001, ** p $<$ 0.01, * p $<$ 0.05, . p $<$ 0.1}\\
\end{tabular}}
\end{table*}

\begin{table*}[!h]
\centering
\caption{\label{tab:visualxspatial_model}Mixed effects logistic regression: AI Model $\times$ Visual PT $\times$ Spatial PT}
\centering
\resizebox{0.9\textwidth}{!}{
\begin{tabular}[t]{lllll}
\toprule
Term & Est./($\chi^2$) & SE/(df) & z & p\\
\midrule
\textbf{A. Model Coefficients} &  &  &  & \\
Intercept & 0.909 & 0.210 & 4.325 & $<$ 0.001***\\
Model: GPT-4o & -0.028 & 0.097 & -0.292 & 0.770\\
Model: o4-mini & 2.199 & 0.138 & 15.918 & $<$ 0.001***\\
Model: o3 & 2.841 & 0.172 & 16.537 & $<$ 0.001***\\
Visual PT: Different & -1.927 & 0.270 & -7.135 & $<$ 0.001***\\
Spatial PT: Different & -5.391 & 0.424 & -12.727 & $<$ 0.001***\\
Format: Image & -0.760 & 0.182 & -4.186 & $<$ 0.001***\\
Model: GPT-4o $\times$ Visual PT: Different & -0.302 & 0.153 & -1.976 & 0.048*\\
Model: o4-mini $\times$ Visual PT: Different & -1.253 & 0.170 & -7.355 & $<$ 0.001***\\
Model: o3 $\times$ Visual PT: Different & -1.404 & 0.198 & -7.080 & $<$ 0.001***\\
Model: GPT-4o $\times$ Spatial PT: Different & 4.011 & 0.356 & 11.271 & $<$ 0.001***\\
Model: o4-mini $\times$ Spatial PT: Different & 1.238 & 0.370 & 3.350 & $<$ 0.001***\\
Model: o3 $\times$ Spatial PT: Different & 5.506 & 0.414 & 13.300 & $<$ 0.001***\\
Visual PT: Different $\times$ Spatial PT: Different & 3.231 & 0.532 & 6.073 & $<$ 0.001***\\
Model: GPT-4o $\times$ Visual PT: Different $\times$ Spatial PT: Different & -2.110 & 0.430 & -4.905 & $<$ 0.001***\\
Model: o4-mini $\times$ Visual PT: Different $\times$ Spatial PT: Different & -1.366 & 0.444 & -3.080 & 0.002**\\
Model: o3 $\times$ Visual PT: Different $\times$ Spatial PT: Different & -4.506 & 0.470 & -9.580 & $<$ 0.001***\\
 &  &  &  & \\
\textbf{B. Type III Wald Tests} &  &  &  & \\
Intercept & 18.703 & 1 &  & $<$ 0.001***\\
Model & 529.906 & 3 &  & $<$ 0.001***\\
Visual PT & 50.906 & 1 &  & $<$ 0.001***\\
Spatial PT & 161.973 & 1 &  & $<$ 0.001***\\
Format & 17.522 & 1 &  & $<$ 0.001***\\
Model $\times$ Visual PT & 85.176 & 3 &  & $<$ 0.001***\\
Model $\times$ Spatial PT & 449.090 & 3 &  & $<$ 0.001***\\
Visual PT $\times$ Spatial PT & 36.886 & 1 &  & $<$ 0.001***\\
Model $\times$ Visual PT $\times$ Spatial PT & 134.516 & 3 &  & $<$ 0.001***\\
\bottomrule
\multicolumn{5}{l}{\textsuperscript{} Panel A: Coefficient estimates with Wald z-tests for individual parameters}\\
\multicolumn{5}{l}{\textsuperscript{} Panel B: Type III Wald $\chi^2$ tests for omnibus effects of factors}\\
\multicolumn{5}{l}{\textsuperscript{} Significance codes: *** p $<$ 0.001, ** p $<$ 0.01, * p $<$ 0.05, . p $<$ 0.1}\\
\multicolumn{5}{l}{\textsuperscript{} Random effect variance (Trial ID): 0.075}\\
\end{tabular}}
\end{table*}

\begin{figure*}[!ht]
    \centering
    \begin{subfigure}[b]{0.8\textwidth}
        \centering
        \includegraphics[width=\textwidth]{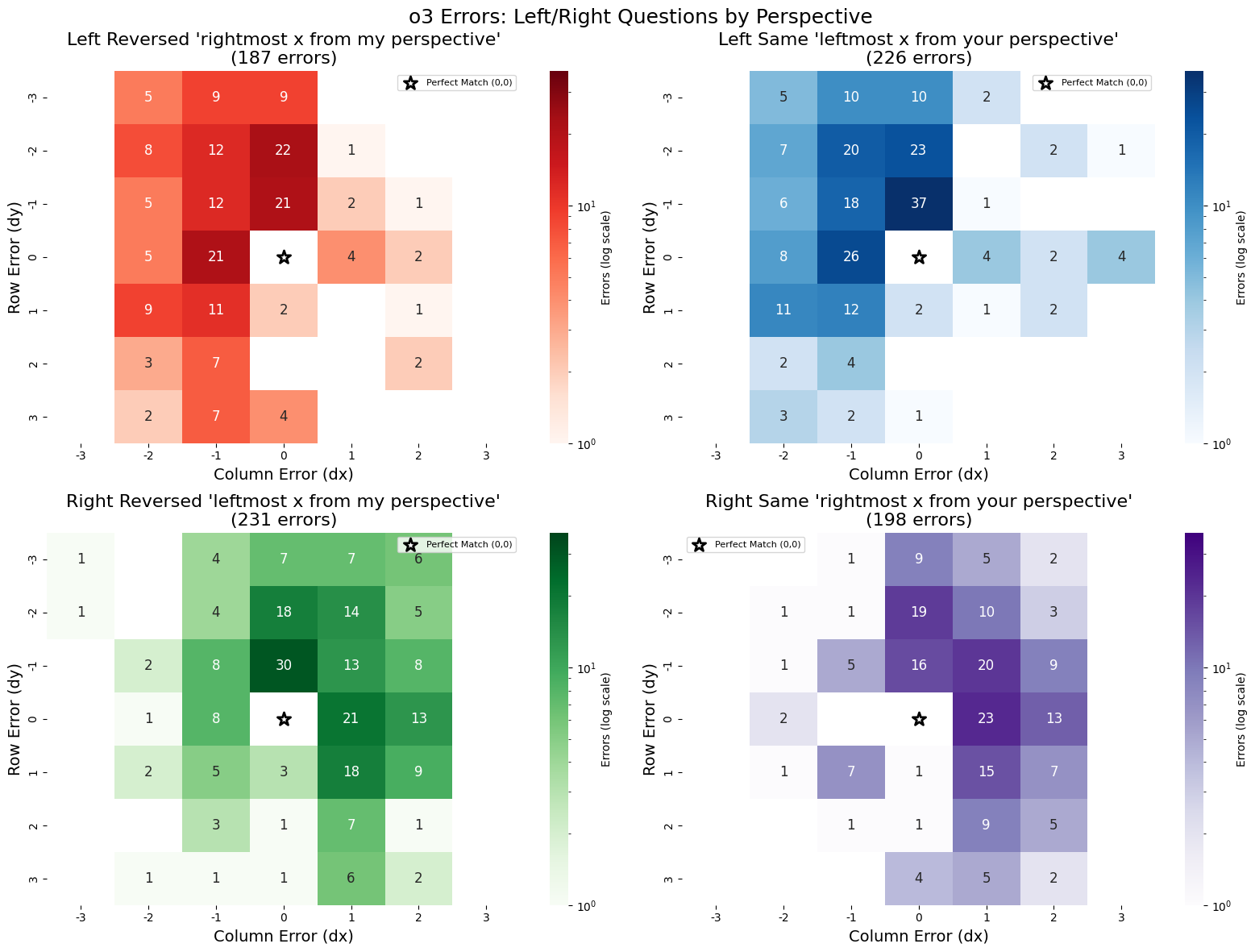}
    \end{subfigure}
    \hfill
    \begin{subfigure}[b]{0.8\textwidth}
        \centering
        \includegraphics[width=\textwidth]{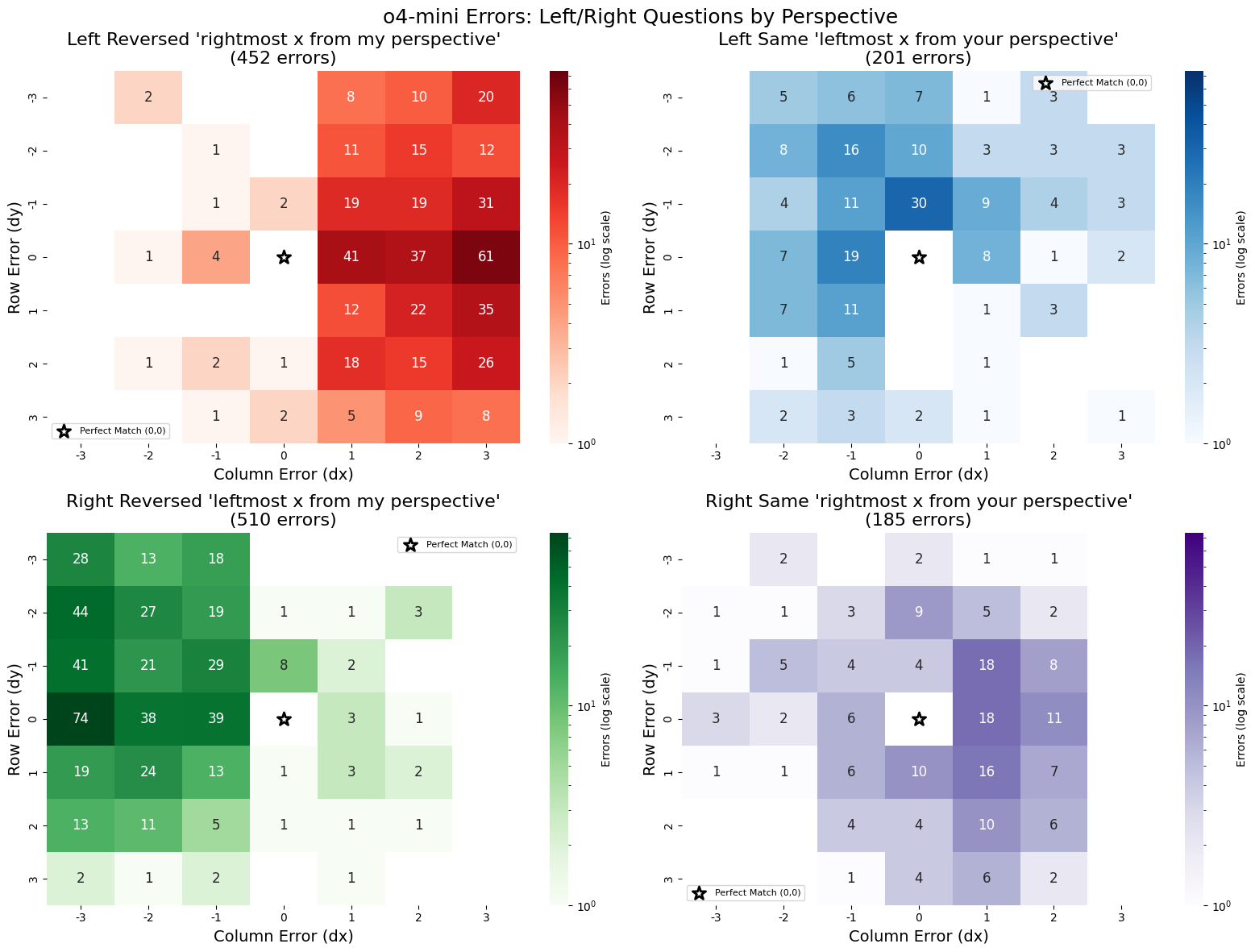}
    \end{subfigure}
    \caption{Relative position of each error made by the models for questions using "leftmost" or "rightmost" selectors split by model and then by perspective.}
    \label{fig:LeftRightErrors}
\end{figure*}

Finally, as the answers for the directors task are coordinates it is possible for us to plot the difference between the correct answer and the answer given for each answer a model got wrong but still gave a valid answer. We find that for samples that used "leftmost" or "rightmost" selectors each model, other than o3, was unable to inhibit its own perspective when the direction was given from the directors point of view. This can be seen clearly in figure \ref{fig:LeftRightErrors} where we can see that for samples of the form "rightmost X from my perspective" models often selected items to the far right of the correct answer rather than realising that the director's right is their left. From this we can also see that for o3 the errors are clustered much closer to the correct answer. 

Additionally, when we compare to the samples using "topmost" or "bottommost" this pattern does not appear with most of the errors occurring to the left and right of the correct answer. This indicates some ability to understand the shared perspective when it does exist.

\begin{figure*}[!ht]
    \centering
    \begin{subfigure}[b]{0.8\textwidth}
        \centering
        \includegraphics[width=\textwidth]{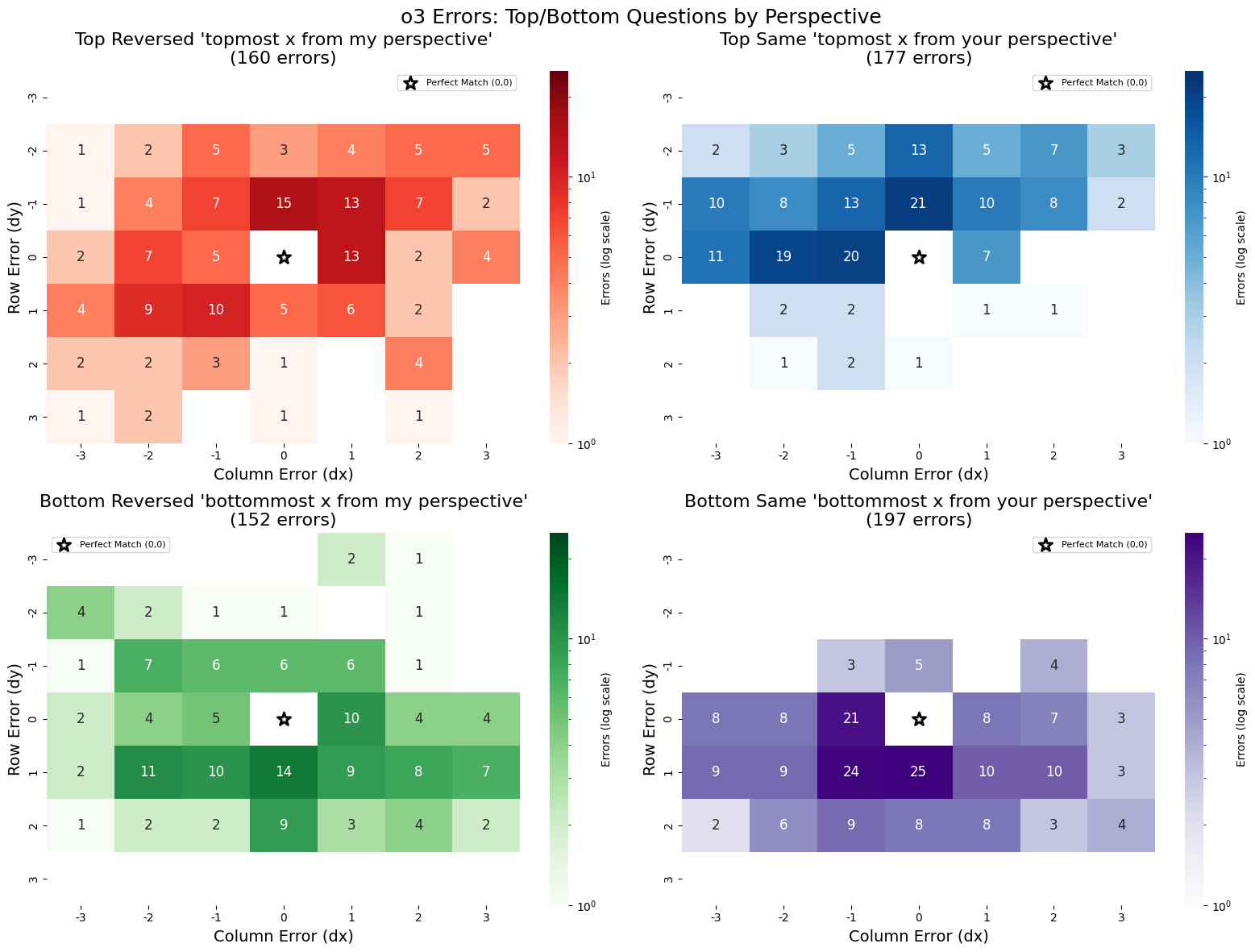}
    \end{subfigure}
    \hfill
    \begin{subfigure}[b]{0.8\textwidth}
        \centering
        \includegraphics[width=\textwidth]{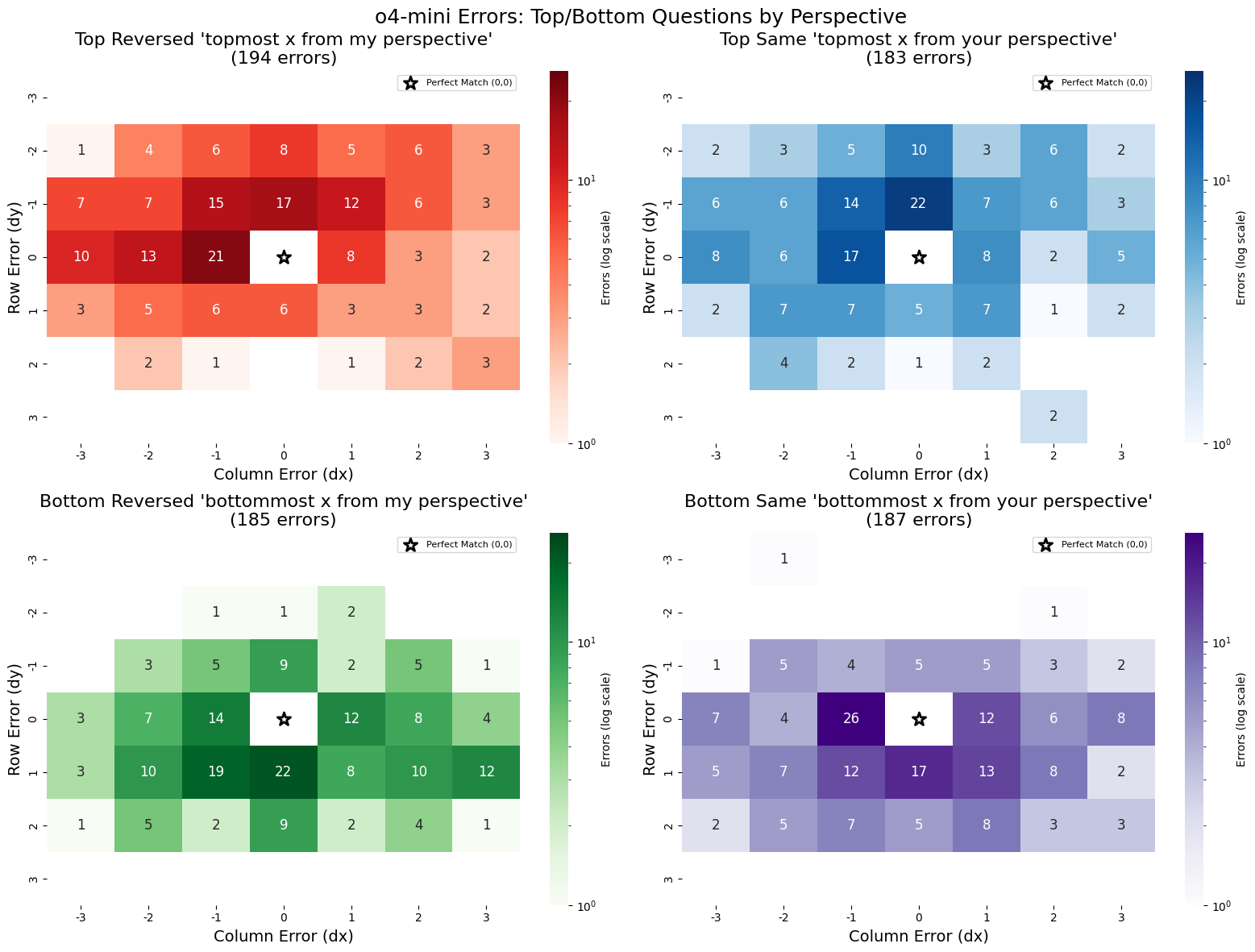}
    \end{subfigure}
    \caption{Relative position of each error made by the models for questions using "topmost" or "bottommost" selectors split by model and then by perspective.}
    \label{fig:TopBottomErrors}
\end{figure*}

\clearpage
\section{Model Details}
For each model we used the default generation parameters provided by OpenAI which are as follows:
\begin{verbatim}
{                                         
      "model": <model_name>,              
      "temperature": 1.0,                 
      "top_p": 1.0,                       
      "frequency_penalty": 0.0,           
      "presence_penalty": 0.0,            
      "reasoning_effort": "medium",       
  } 
\end{verbatim}

The reasoning effort parameter was only set for models o4-mini and o3 as they were the only reasoning models used. Some models are updated over the course of their deployment and when we refer to a model such as gpt-4o we are refering to the specific checkpoint we used which is as follows:

\begin{itemize}
    \item o3-2025-04-16
    \item o4-mini-2025-04-16
    \item gpt-4o-2024-11-20
    \item gpt-4o-mini-2024-07-18
\end{itemize}

\end{document}